\documentclass[11pt]{scrartcl}

\usepackage[utf8]{inputenc}
\usepackage[T1]{fontenc} 
\usepackage[authoryear,round]{natbib}
\usepackage{amsfonts, amsmath, amssymb, amsthm}
\usepackage{bm}
\usepackage{thmtools}
\usepackage{graphicx}
\usepackage{xcolor} 

\usepackage[automark]{scrlayer-scrpage}
\clearpairofpagestyles
\ohead{\pagemark}
\ihead{\headmark}

\usepackage{booktabs, tabularx, multirow, adjustbox}
\usepackage{float, subfig}

\usepackage{algorithm2e}
\usepackage{fancyvrb}
\VerbatimFootnotes
\usepackage{enumitem} 
\usepackage[all]{xy}
\usepackage{ifthen}

\usepackage[a4paper,
            top=2cm,
            bottom=2cm,
            textwidth=15cm,
            footskip=1.5cm,
            headsep=0.5cm]{geometry}
\usepackage{setspace}
\usepackage{authblk}
\usepackage{courier}

\newtheorem{definition}{Definition}[section]
\newboolean{showSolutions}
\setboolean{showSolutions}{true}
\newcommand{\solutioncmd}[2]{{{#1}}{{#2}}}
\newcommand{\solution}[2]{\ifthenelse {\boolean{showSolutions}} {\solutioncmd{#1}}{\solutioncmd{#2}}}

\usepackage[colorlinks=true, urlcolor=blue, citecolor=blue, linkcolor=blue]{hyperref}
\title{Conformal Prediction for Compositional Data}
\author[1, 2]{Lucas P. Amaral}
\author[1,2]{Luben M. C. Cabezas}
\author[1]{Thiago R. Ramos}
\author[1]{Gustavo H. A. Pereira}
\affil[1]{Department of Statistics, Federal University of São Carlos}
\affil[2]{Institute of Mathematics and Computer Science, University of São Paulo}
\date{}

\begin{document}
\maketitle

\begin{abstract}
\noindent Dirichlet regression models are suitable for compositional data, in which the response variable represents proportions that sum to one. However, there are still no well-established methods for constructing valid prediction sets in this context, especially considering the geometry of the compositional space. In this work, we investigate conformal prediction-based strategies for constructing valid predictive regions in Dirichlet regression models. We evaluate three distinct approaches: a method based on quantile residuals, an approximate construction of highest density regions (HDR), and an adaptation of the approximate HDR using grid-based discretization over the simplex. The performance of the methods was analyzed through simulation studies under different scenarios, varying the model complexity, response dimensionality, and covariate structure. The results indicated that the HDR approximation approach exhibits good robustness in terms of coverage, while the grid discretization proved effective in reducing overcoverage and the area of the prediction region compared to the original method. The quantile method provided larger prediction regions compared to the grid method, while maintaining adequate coverage. The methodologies were also applied to two real datasets: one concerning sleep stages and another on biomass allocation in plants. In both cases, the proposed methods demonstrated practical feasibility and produced coherent interpretations within the compositional space. Finally, we discuss possible extensions of this work.\end{abstract}


\section{Introduction}
\newcommand{\comando}[1]{\textbf{\textbackslash#1}}

Compositional data (CoDa) are widely used across many fields whenever the goal is to model the parts of a whole. Typical examples include ecology, to quantify species shares in an ecosystem \citep{greenacre2021compositional}, medicine, to measure the concentrations of different cell types in a patient’s blood \citep{zhang2025compositional}, among others.
A compositional observation is a vector $\mathbf{y}=(y_1,\dots,y_D)^\top$ of positive components such that $\sum_{j=1}^{D}y_j=1$. The set
\[
\Delta^{D}=\{(y_1,\dots,y_D)\in\mathbb{R}^{D}:\ \sum_{j=1}^{D}y_j=1,\ y_j>0,\ j=1,\dots,D\}
\]
is denoted the $(D-1)$-simplex.
CoDa can be analyzed in several ways, but modeling is more used due to the simplex constraints. Over the years, however, substantial progress has been made (see \cite{alenazi2023review}). The simplex is the natural sample space for these data \citep{aitchison1982statistical}, and the Dirichlet family is a common probabilistic model on this space \citep{barndorff1991some}. In regression contexts, Dirichlet regression models \citep{hijazi2009modelling} are frequently employed and can be viewed as generalizations of beta regression models \citep{ferrari2004beta}.
Parametric modeling often emphasizes the interpretability of regression coefficients and the verification of modeling assumptions, as these directly affect the goodness of fit. 

Nevertheless, in many practical scenarios, the primary focus lies in the prediction of new observations and their respective predictive intervals. When response variables are constrained to the unit interval $(0,1)$, as is the case for proportions or fractions, traditional approaches based on normality assumptions are often inadequate. In these cases, there are established methodologies that rely on resampling techniques (bootstrap) and employ the residuals as surrogates to estimate the prediction error \citep{espinheira2014bootstrap,cribari2021resampling}. However, such methods typically rely on asymptotic theory and may require large samples to achieve accurate coverage. Moreover, standard bootstrap procedures may be ill–suited under heteroscedasticity, as this can undermine exchangeability and lead to invalid inference and predictions. Recently, \cite{wu2025conformalized} proposed conformalized methods for bounded outcomes within the broader Conformal Prediction (CP) framework \citep{vovk2005algorithmic,shafer2008tutorial,papadopoulos2008normalized, lei2018distribution}, offering a machine-learning-based approach to construct prediction intervals with finite–sample guarantees, including marginal and asymptotic conditional validity. 

Inspired by this successful adaptation to bounded data, we leverage the fundamental principles of CP\citep{vovk2005algorithmic,shafer2008tutorial,lei2018distribution}. CP is a distribution–free and model–agnostic framework highly valued for providing marginal coverage guarantees in finite samples under the mild assumption of exchangeability. This assumption is often reasonable in various data settings and allows CP to be coupled with a wide range of statistical models and learning algorithms \citep{angelopoulos2021gentle}. The core ingredient of this framework is the nonconformity score, a function that quantifies how atypical a data point is relative to a model trained on the remaining data.

Numerous extensions for general regression models have been studied \citep{tibshirani2019conformal}, along with proposals for constructing nonconformity scores in different contexts \citep{kato2023review}. A persistent challenge in modeling is heteroscedasticity, i.e., when the response variance varies with covariates. In Dirichlet regression, the mean and variance are intrinsically linked; thus, dispersion is naturally accommodated by the model. To address heteroscedasticity within conformal prediction, several adaptations have been developed, including conformalized quantile regression \citep{romano2019conformalized}, normalized conformal prediction \citep{papadopoulos2008normalized,lei2018distribution,kato2023review}, and Mondrian conformal prediction \citep{bostrom2020mondrian,cabezas2025regression,dewolf2025conditional}.

A model–based extension of conformal prediction tailored to distributional regression was introduced by \cite{chernozhukov2021distributional}. Their approach uses the probability integral transform (PIT) to define nonconformity measures from an estimated conditional distribution of the response, yielding asymptotically valid prediction intervals under heteroscedasticity. Some related works, \cite{tibshirani2019conformal,barber2023conformal} propose conformal prediction algorithms based on weighted quantiles that deliver finite–sample guarantees in broad, non-exchangeable scenarios.

Inspired by these ideas, this work proposes conformal prediction methods for compositional data. The paper is organized as follows. Section~\ref{sec:dirichlet-regression} presents two parameterizations of the Dirichlet distribution and the corresponding regression specification. Section~\ref{sec:predictive-intervals} details our approach for constructing conformal prediction sets for Dirichlet regression using (i) randomized quantile residuals \citep{dunn1996randomized} of the marginal components, (ii) an approximation to the highest density region (HDR) that contains the exact prediction set, and (iii) a grid search within the resulting polytope to mitigate overcoverage and overly wide intervals. Section~\ref{sec:simulation-studies} describes the simulation design and reports results for the three predictive methods, evaluating empirical coverage, interval widths, and runtime. Section~\ref{sec:application} presents two real data applications. Section~\ref{sec:conclusions} concludes with a general discussion and avenues for future work. R code and data can be found in the GitHub repository https://github.com/LucAmaralDS/CP-dirichlet.

\section{Regression models for compositional data}
\label{sec:dirichlet-regression}
Consider a regression setting for CoDa, where the $D$-part response vector $\mathbf{Y}_i = (Y_{i1}, \dots, Y_{iD})^\top$ for the $i$-th observation, $i = 1, \dots, n$, belongs to the D-dimensional simplex, $\mathbf{Y}_i \in \Delta^{D}$. 
In many CoDa regression problems, we assume that the response $\mathbf{Y}_i$ follows a Dirichlet distribution with shape parameters $\boldsymbol{\lambda}_i = (\lambda_{i1}, \dots, \lambda_{iD})^\top$, where $\lambda_{ij} > 0$ for all $j=1, \dots, D$. The probability density function (PDF) for a general compositional vector $\mathbf{Y}$ following $\text{Dirichlet}(\boldsymbol{\lambda})$ is given by:$$f(\mathbf{y}; \boldsymbol{\lambda}) \;=\; \frac{\Gamma\!\big(\sum_{j=1}^{D}\lambda_j\big)}{\prod_{j=1}^{D}\Gamma(\lambda_j)} \prod_{j=1}^D y_j^{\lambda_j - 1}, \quad \mathbf{y} \in \Delta_D$$where $\Gamma(\cdot)$ is the Gamma function.

\noindent Let $\lambda_0 = \sum_{t=1}^D \lambda_t$. For the Dirichlet distribution, the mean and variance are, respectively,
\begin{equation*}
    \mathbb{E}[Y_j] \;=\; \frac{\lambda_j}{\lambda_0}
    \qquad \text{and} \qquad
    \mathrm{Var}(Y_j) \;=\; \frac{\lambda_j(\lambda_0 - \lambda_j)}{\lambda_0^2(\lambda_0 + 1)}.
\end{equation*}

 For regression problems, where the response $\mathbf{Y}_i$ is linked to a vector of covariates $\mathbf{X}_i$, it is common to adopt a Dirichlet parameterization that models the mean of the response components ($\mu_j$) together with a precision parameter ($\phi$), yielding more direct interpretability of the estimated coefficients.
 We define the reparameterization by letting $\phi = \sum_{j=1}^D \lambda_j$ and $\mu_j = \lambda_j/\phi$, for $j = 1, \dots, D$, with $\sum_{j=1}^D \mu_j = 1$. The density function under this alternative mean-precision parameterization is given by:
 \begin{equation}
     f(\mathbf{y}; \mu_1,\dots,\mu_D,\phi) \;=\;
 \frac{\Gamma(\phi)}{\prod_{j=1}^D \Gamma(\mu_j \phi)}
 \prod_{j=1}^D y_j^{\phi \mu_j - 1} \label{dirichlet-rep}
 \end{equation} where $0 < \mu_j < 1$ for all $j$ and $\phi > 0$. Under this setup, the expected value and variance of the $j$-th component are, respectively:$$\mathbb{E}[Y_j] = \mu_j \quad \text{and} \quad \mathrm{Var}[Y_j] \;=\; \frac{\mu_j(1-\mu_j)}{\phi+1} = \frac{V(\mu_j)}{\phi+1}.$$The term $\phi$ is thus a precision parameter, as larger values lead to smaller variance.

Let $Y_1, \ldots, Y_n$ and 
$\mathbf{Y}_i = (Y_{i1}, \ldots, Y_{iD})^\top$ be random variables distributed 
according to a Dirichlet distribution as in Eq. \eqref{dirichlet-rep}. The systematic components of 
the logistic Dirichlet regression model are given by\begin{equation}
\label{regressao-especificada}
\begin{aligned}
\log\left(\frac{\mu_{ij}}{\mu_{i1}}\right)
  &= x_{ij1}\beta_{j1} + x_{ij2}\beta_{j2} + \cdots + x_{ij p_j}\beta_{j p_j},
  && 2 \le j \le D, \\
\log(\phi_i)
  &= d_{i1}\gamma_1 + d_{i2}\gamma_2 + \cdots + d_{i p_\phi}\gamma_{p_\phi},
\end{aligned}
\end{equation} $(x_{ij1}, x_{ij2}, \ldots, x_{ij p_j}, d_{i1}, d_{i2}, \ldots, d_{i p_\phi})^\top$ 
are the covariates and $(\beta_{j1}, \beta_{j2}, \ldots, \beta_{j p_j}, 
\gamma_1, \gamma_2, \ldots, \gamma_{p_\phi})^\top$ unknown parameters. We adopt a logit link for the mean vector and a logarithmic link for the precision parameter, in line with the conventional parametrization of multinomial logistic regression \citep{hosmer2013applied}. This specification is particularly convenient because it yields regression coefficients with a transparent interpretation and automatically restricts the precision parameter to the positive real line, satisfying the distributional constraint $\phi > 0$.
The parameters of model \eqref{regressao-especificada} are estimated by maximum likelihood, and all empirical results reported here are obtained using the \texttt{DirichletReg} package \citep{maier2014dirichletreg} in \textsf{R}.

\section{Conformal prediction for compositional data regression models}
\label{sec:predictive-intervals}

A common practice in regression analysis is to use a fitted model to predict out-of-sample or missing response values. While a point prediction can be readily obtained from the fitted model, it is often desirable to accompany it with a prediction region at a prescribed coverage level. In this section, we present our proposals for constructing conformal prediction regions in Dirichlet regression models. Section~\ref{secao-conformal} introduces the conformal prediction framework and the specific algorithm used throughout, namely split conformal prediction. In Section~\ref{secao-conformal-resq}, we construct prediction regions using quantile residuals \citep{dunn1996randomized} as nonconformity scores. Section~\ref{sec:hdr-approx} instead uses the negative log-likelihood as the nonconformity score and inverts it to obtain a highest-density prediction region, which we approximate by coordinate-floor. The associated convex optimization formulation underlying this approximation is presented in Section~\ref{sec:hdr-approx-optim}.

\subsection{Conformal prediction} \label{secao-conformal}
Recently, conformal prediction (CP) methods have been widely employed for constructing prediction regions under weak assumptions \cite{vovk2005algorithmic,shafer2008tutorial}. Given an exchangeable sample of pairs $(X_i, Y_i)$, $i = 1,\dots,n$, conformal prediction is used to produce a predictive set $\mathcal{C}(\cdot)$ for a new covariate vector $X_{n+1}$ with an unknown response $Y_{n+1}$, such that
$\mathbb{P}\bigl(Y_{n+1} \in \mathcal{C}(X_{n+1})\bigr) \geq 1 - \alpha.$
The literature offers several algorithms for constructing these predictive sets, including Full Conformal Prediction (FCP) \cite{vovk2005algorithmic}, Split Conformal Prediction (SCP) \cite{papadopoulos2008normalized,vovk2012conditional,lei2018distribution}, and Jackknife+ \cite{barber2021predictive}.

In this work, we are interested in predictions for compositional data in large samples. In such cases, the SCP method is more convenient and much more commonly used due to its significantly lower computational cost compared to the others. Therefore, we will consider the SCP method in this study. This method partitions the available data into a training set ($\mathcal{D}{\text{train}}$) and a calibration set ($\mathcal{D}{\text{cal}}$). While the model is fitted only on $\mathcal{D}_{\text{train}}$, this split implies a statistical power trade-off, potentially leading to wider predictive sets. However, when the sample is small, SCP becomes infeasible precisely because it requires splitting the data into training and calibration sets.

The central component of CP is the nonconformity score, denoted by $s(\mathbf{X}, Y)$, defined as a function $s: \mathcal{X} \times \mathcal{Y} \to \mathbb{R}$, where $\mathcal{X} \subseteq \mathbb{R}^p$ and $\mathcal{Y} \subseteq \mathbb{R}$ (or in the compositional case, $\mathcal{Y} \subseteq \Delta^D$), which quantifies how atypical a new observation $(\mathbf{X}_{n+1}, Y_{n+1})$ is. For a significance level $\alpha \in (0, 1)$, the $(1-\alpha)$-level predictive region of SCP, $\mathcal{C}(\mathbf{X}_{n+1})$, is generally defined over the response domain $\mathcal{Y}$ as:
$$
\mathcal{C}(\mathbf{x}_{n+1}) = \left\{ y_{n+1} \in \mathcal{Y} : s(\mathbf{x}_{n+1}, y_{n+1}) \le q_{1-\alpha} \right\},
$$
where $q_{\alpha}$ is the empirical quantile of order $\lceil (1-\alpha)(n_{\text{cal}} + 1) \rceil$ of the nonconformity scores computed on the calibration set $\mathcal{D}_{\text{cal}}$, and $n_{\text{cal}}$ is the size of the calibration set. This region provides a marginal coverage guarantee of at least $1-\alpha$ in finite samples under the exchangeability assumption.

In CP approaches, the choice of $s(\mathbf{x}, y)$ is critical, as it directly influences the shape and informativeness of the resulting predictive regions \cite{angelopoulos2021gentle,Izbicki2025}. In regression for continuous, unbounded univariate responses, a common choice for $s(\mathbf{x}, y)$ is the ordinary absolute residual, $|y-\widehat{\mathbb{E}}[Y|\mathbf{x}]|$ \cite{lei2018distribution}, where $\widehat{\mathbb{E}}[Y|\mathbf{x}]$ is the estimated regression function fitted on the training data.

However, the compositional nature of the data and the use of the Dirichlet distribution require scores that naturally account for the distribution's properties and intrinsic geometry. For this reason, we focus on constructing predictive regions over the simplex $\Delta^D$. In particular, we choose to use nonconformity scores based on the quantile residual \cite{dunn1996randomized} and the negative log-likelihood, which are detailed in the following sections. The complete procedure for constructing the predictive regions using these specific scores is presented in Algorithms \ref{alg:dirichlet_split_conformal_en} and \ref{alg:dirichlet_triangular_hdr_en}, which appear in \ref{apendiceA}.

\subsection{Predictive sets with quantile residuals} \label{secao-conformal-resq}
The first CP method proposed in this work uses the quantile residual. The quantile residual, proposed by \cite{dunn1996randomized}, is a simple-to-use measure that is applicable to various regression models. When model parameters are consistently estimated, this residual is asymptotically standard normally distributed. Moreover, studies indicate that even with small samples, its distribution is often well approximated by the standard normal in different regression models \citep{pereira2019quantile, lemonte2019residuals, feng2020comparison}. For this reason, the quantile residual has been widely used for regression model diagnostics.

In the case of Dirichlet regression, our first proposal is to use split conformal with the quantile residual, taking advantage of the fact that the marginal distributions of each component of a Dirichlet vector follow a beta distribution. Since we are using the alternative parameterization of the Dirichlet distribution, its marginals are expressed in terms of the beta distribution parameterization proposed by \cite{ferrari2004beta}. In particular, for each component $j=1,\dots,D$, conditionally on the covariates $\mathbf{X}=\mathbf{x}$, we have:

\[
Y_j \mid \mathbf{X}=\mathbf{x} \sim \text{Beta}\big(\mu_j(\mathbf{x})\,,\phi(\mathbf{x})),
\]
where $\mu_j(\mathbf{x}) \in (0,1)$ represents the conditional mean of component $j$ and $\phi(\mathbf{x}) > 0$ is the precision parameter of the Dirichlet regression model.

Given the fitted model, we obtain the estimates $\widehat{\mu}_j(\mathbf{x})$ and $\widehat{\phi}(\mathbf{x})$, which characterize the conditional distribution of each marginal component. Based on this parametrization, the nonconformity score is defined using the quantile residual. For observation $i$ and component $j$, the quantile residual is given by
\begin{equation}
r_{ij}^q
=
\Phi^{-1}\!\left\{
F(y_{ij};\widehat{\mu}_{ij},\widehat{\phi}_i)
\right\},
\label{residuo-quantilico-mu-phi}
\end{equation}
where $\widehat{\mu}_{ij}=\widehat{\mu}_j(\mathbf{x}_i)$ and $\widehat{\phi}_i=\widehat{\phi}(\mathbf{x}_i)$. Here, $\Phi(\cdot)$ denotes the cumulative distribution function (CDF) of the standard normal distribution and $F(\cdot)$ denotes the CDF of the beta distribution corresponding to component $j$.

However, it is necessary to construct a prediction region in which all components are simultaneously within their respective intervals. To achieve this, several scores could be used, such as Mahalanobis ellipses \cite{ghorbani2019mahalanobis}, but in such cases, it would not be possible to invert the terms and construct the intervals without resorting to grid search. In light of these considerations, we propose the following nonconformity score:
\begin{equation}
s(\mathbf{x},\mathbf{y}) = \text{max}_j|r_{j}^q|,
\end{equation}
where $r_{ij}^q$ is given by \eqref{residuo-quantilico-mu-phi}. In Algorithm \ref{alg:dirichlet_split_conformal_en}, we describe the step-by-step procedure for implementing this method.

Note that the predictive set is, by definition, $\mathcal{C}{(\mathbf{x}_{n+1})} = \{\mathbf{y} \in \Delta^{D}: s(\mathbf{x},\mathbf{y}) \leq q_{1-\alpha}\}$. Thus, by using the maximum of the absolute value of the quantile residual from the $j$-th component, we apply a global calibration, since $\text{max}_j|r_{j}^q| \leq q_{1-\alpha}$ ensures that all components with lower scores in magnitude satisfy $|r_{j}^q| \leq q_{1-\alpha}$. Analyzing component by component, this is equivalent to:
\begin{equation*}
-q_{1-\alpha} \leq r_{j}^q \leq q_{1-\alpha} \ \ \ \Leftrightarrow \ \ \ \ \Phi(-q_{1-\alpha}) \ \leq F_j \leq \ \Phi(q_{1-\alpha}),
\end{equation*}
because the standard normal CDF $\Phi(\cdot)$ is a monotonic function, meaning it preserves inequality. So, let $p_{\text{inf}} = \Phi(-q_{1-\alpha})$ and $p_{\text{sup}} = \Phi(q_{1-\alpha})$, we obtain closed marginal intervals in the original response scale, given by
\begin{equation*}
    I_j = [ F^{-1}_j(p_{\text{inf}}), \ F^{-1}_j(p_{\text{sup}})].
\end{equation*}
Therefore,
\begin{equation*}
\mathcal{C}(\mathbf{x}_{n+1}) = \{\mathbf{y} \in \Delta^{D}: y_j \in I_j(\mathbf{X}_{
n+1}) \ \forall \ j \}.
\end{equation*}


\subsection{Highest Density Regions (HDR)}

A natural way to summarize the uncertainty associated with a multivariate random variable is through the construction of confidence regions that, for a given probability level, occupy the smallest possible volume in the sample space. Such sets are called highest density regions or, in English, \textit{highest density regions} (HDR).

\begin{definition}[Highest density region; \cite{hyndman1996computing}]
    Let $f$ be the probability density function of a continuous and possibly multivariate random variable $\mathbf{Y} \in \mathbb{R}^d$; the $100(1-\alpha)\%$ HDR is defined as the subset $\mathcal{C}(f_\alpha)$ of the sample space of $\mathbf{y}$ such that:
    \begin{equation}
        \mathcal{C}(f_\alpha) = \{\mathbf{y} : f(\mathbf{y}) \ge f_\alpha\},
         \label{eq:hdr-def}
    \end{equation}
    where $f_\alpha$ is the largest constant such that $P(\mathbf{Y} \in \mathcal{C}(f_\alpha)) \ge 1 - \alpha$, with $\alpha \in (0, 1)$.
\end{definition}

\cite{hyndman1996computing} highlights a fundamental property of HDRs: among all regions with probability coverage of at least $1-\alpha$, the region $\mathcal{C}$ defined in \eqref{eq:hdr-def} has the smallest volume with respect to the Lebesgue measure. Geometrically, the boundary of an HDR corresponds to a level set of the density $f(\mathbf{y})$.

This property makes HDRs particularly attractive for multimodal or highly skewed distributions, as is often the case for the Dirichlet distribution on the simplex. Unlike regions based on central moments or symmetric intervals, the HDR adapts to the geometry of the density, capturing probability where it is most concentrated.

\begin{figure}[!htb]
    \centering
    \includegraphics[scale = 0.5]{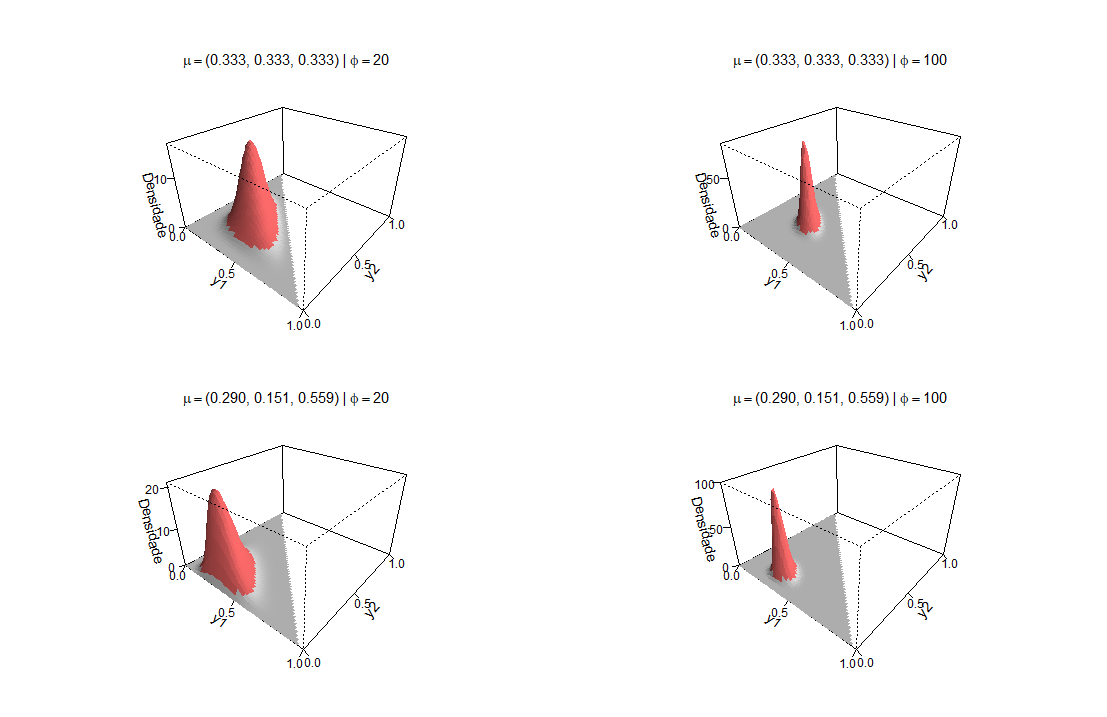}
    \caption{Graphical representation of the HDR in 3D for three components and different values of the Dirichlet distribution parameters.}
    \label{fig:HDR-3D}
\end{figure}

Figure \ref{fig:HDR-3D} presents Dirichlet density surfaces on the simplex for two choices of $\mu$ and two values of $\phi$. The red region indicates the HDR for a fixed probability level $1-\alpha$, with $\alpha = 0.10$. Comparing $\phi=20$ with $\phi=100$, it can be observed that increasing $\phi$ makes the distribution much more concentrated, the peak becomes higher and narrower, and the HDR shrinks, occupying a smaller area of the simplex. This reflects lower dispersion of the compositions around the center.

One can note that $\bm{\mu}$ controls the location of the highest probability and the positioning of the HDR. In the symmetric case, the mass concentrates away from the extremes of the simplex, yielding a centered HDR; in the asymmetric case, the peak shifts toward the dominant component (largest $\mu_j$), and the HDR approaches the region of the simplex where this component is higher. Thus, $\bm{\mu}$ determines where the typical compositions lie, while $\phi$ determines how tight the concentration is around them.

\begin{figure}[!htb]
    \centering
    \includegraphics[scale = 0.5]{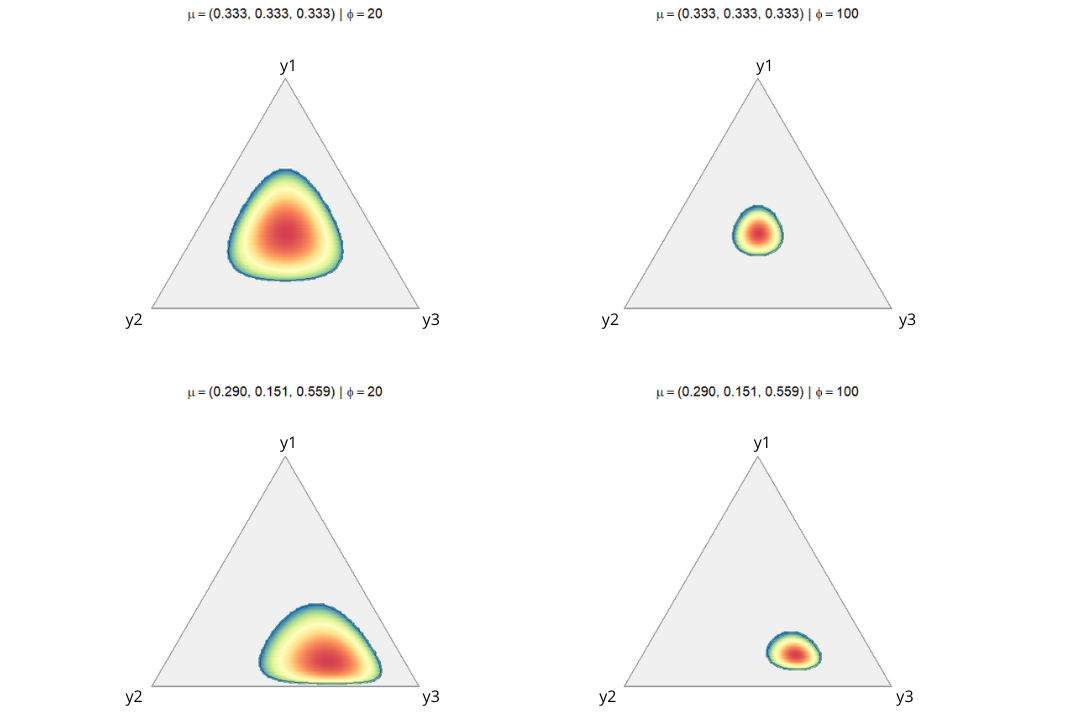}
    \caption{Ternary plot representation of the HDR for three components and different values of the Dirichlet distribution parameters.}
    \label{fig:HDR-ternario}
\end{figure}

For compositional data, a common practice for visualizing results when there are three response components is to use a ternary plot. In a ternary diagram, the position of the compositions indicates clear patterns: concentration near a vertex suggests dominance of one component; concentration along an edge indicates joint dominance of the two components forming that edge; and points near the barycenter indicate approximately equal proportions.

In the ternary plot with HDR shown in Figure \ref{fig:HDR-ternario}, the core (more intense colors) indicates the region of highest density, while the contour delineates the HDR, the set of points in the simplex whose density exceeds a chosen threshold to contain $1-\alpha$ of the probability. Outside this contour, the density is lower and the remaining probability mass is $\alpha$.

In the context of conformal prediction, the connection with HDRs is established through the choice of the nonconformity function. If we define the score as the negative log-likelihood, $s(\mathbf{x},\mathbf{y}) = -\log f(\mathbf{y}|\mathbf{x})$, then the prediction sets generated by the conformal method will consist of those points $\mathbf{y}$ for which the density $f(\mathbf{y}|\mathbf{x})$ exceeds a data-calibrated threshold, thus yielding valid estimators of highest density regions.

\subsection{Conformal prediction on the simplex with highest density region approximation}
\label{sec:hdr-approx}
As discussed in the previous section, the highest density region (HDR) lies at the core of the method and contains the points with the highest probability. Hence, it defines the predictive set. We propose to approximate the HDR using a floor-polytope that encloses it. It is worth noting that a polytope is simply a generalization of the terms polygon and polyhedron to any dimension. For example, in two dimensions a polytope is a polygon, in three it is a polyhedron.

We define a floor-polytope as the subset of the simplex obtained by imposing lower bounds $y_i \geq \tau_i$. That is, 
\[
\mathcal{T}(\bm{\tau}) = \{ \mathbf{y} \in \Delta^D: y_i \geq \tau_i, \ i = 1,\dots,D \},
\]
which corresponds to a truncated simplex and will be chosen to contain the exact region $\mathcal{C}_{\text{exact}}(\mathbf{x})$.

Consider the Dirichlet regression model defined in \eqref{regressao-especificada}. Suppose that given $\mathbf{x}$, the model provides parameters for a Dirichlet distribution with density given by \eqref{dirichlet-rep}. For simplicity, we denote $\mu_j(\mathbf{x}) = \mu_j$ and $\phi(\mathbf{x}) = \phi$, although they depend on $\mathbf{x}$, as the model estimates will be used. We define the nonconformity score as
\[
s(\mathbf{x},\mathbf{y}) = -\log f(\mathbf{y}\mid \mathbf{x}) 
= -\log \Gamma(\phi) + \sum_{j=1}^{D}\log\Gamma(\mu_j\phi) - \sum_{j=1}^{D}(\mu_j\phi - 1)\log y_j,
\]
since $s = -\log f$ is strictly decreasing in $f$, the set $\{ y: \ s(\mathbf{x},\mathbf{y}) \leq c \}$ is a level set of the conditional density (HDR) given $\mathbf{x}$.

In the calibration set, we compute $s_i=s(\mathbf{x}_i,\mathbf{y}_i)$ and define the threshold as the quantile  
\[
q_{1-\alpha}\;=\;\text{order-}k\text{ of }\{s_i\},\qquad
k=\big\lceil(1-\alpha)\,(n_{\mathrm{cal}}+1)\big\rceil.
\]

After obtaining the conformal quantile, we can invert the score so that we isolate $\sum_{j=1}^{D}(\mu_j\phi - 1)\log y_j$, which contains the response components of interest for the approximation. For the test point $\mathbf{x}_{n+1}$, let $\bm{\mu}=\bm{\mu}(\mathbf{x}_{n+1})$ and $\phi=\phi(\mathbf{x}_{n+1})$, then
\[
t_{1-\alpha} = -q_{1-\alpha} - \log \Gamma(\phi) + \sum_{j=1}^{D}\log\Gamma(\mu_j\phi),
\qquad
w_j = \phi(\mathbf{x}_{n+1})\mu_j(\mathbf{x}_{n+1})-1,
\]
where $t_{1-\alpha}$ depends on $q_{1-\alpha}$, and the exact region is
\begin{equation}
\mathcal{C}_{\text{exact}}(\mathbf{X}_{n+1}) = \Bigl\{ \mathbf{y}\in\Delta^D : \sum_{j=1}^D w_j \log y_j \ge t_{1-\alpha} \Bigr\}.
\label{score-hdr}
\end{equation}

Unlike quantile-residual-based scores, the log-likelihood-based nonconformity score does not allow for an analytical inversion in terms of the response vector. To overcome this, we approximate the HDR with a polytope derived from component-wise lower bounds. The approximation process is detailed in the next section.

\subsection{HDR approximation: optimization formulation}
\label{sec:hdr-approx-optim}
Fix $\mathbf{x}_{n+1}$, $w_j=\phi\mu_j-1$ and $W=\sum_{j=1}^D w_j$. The exact set for the score $s(\mathbf{x},\mathbf{y})=-\log f(\mathbf{y}\mid \mathbf{x})$ is
\[
\mathcal{C}_{\mathrm{exact}}(\mathbf{X}_{n+1}) = \Bigl\{\,\mathbf{y}\in\Delta^D:\ \sum_{j=1}^D w_j\log y_j\ \ge\ t_{1-\alpha} \Bigr\}, \qquad
t_{1-\alpha}=-q_{1-\alpha}-\log \Gamma(\phi) + \sum_{j=1}^D\log \Gamma(\phi\mu_j).
\]

If $w_j>0$ for all $j$, the function
\[
g(\mathbf{y})=\sum_{j=1}^D w_j\log y_j
\]
is concave in the interior of the simplex, since $\log(\cdot)$ is concave and the weighted sum with positive weights preserves concavity. Thus, the exact region is a convex set defined by
\[
\mathcal{C}_{\mathrm{exact}}(\mathbf{x}_{n+1})=\{\mathbf{y}\in\Delta^{D}:\ g(\mathbf{y})\ge t_{1-\alpha}\},
\]
considering $\mathbf{y}$ in the interior of the simplex $(y_j > 0)$, because the score involves $\log y_j$.

The threshold $g(\mathbf{y})\ge t_{1-\alpha}$ tends to exclude points near the simplex boundaries where some component is very small. In fact, since $\log y_j \to -\infty$ as $y_j\downarrow 0$ and $w_j>0$, very small values of $y_j$ greatly reduce $g(\mathbf{y})$, violating the inequality.

Based on this, we construct a simple geometric approximation by imposing lower bounds $y_i\ge \tau_i$. Geometrically, this shifts (truncates) each face $y_i=0$ to the parallel face $y_i=\tau_i$, producing a truncated simplex that can be chosen to contain the exact region.

For each coordinate $i$, the minimum floor $\tau_i$ is given by solving the convex problem
\[
\min_{\mathbf{y}\in\mathbb{R}^D}\ y_i
\quad\text{subject to}\quad
\sum_{j=1}^D y_j=1,\quad
\sum_{j=1}^D w_j\log y_j \ge t_{1-\alpha},\quad
y_j\ge 0\ \ (j=1,\dots,D).
\tag{P$_i$}
\]

We define $\tau_i$ as the smallest admissible value of coordinate $y_i$ inside $\mathcal{C}_{\mathrm{exact}}(\mathbf{x}_{n+1})$. Thus, for all $\mathbf{y}\in\mathcal{C}_{\mathrm{exact}}(\mathbf{x}_{n+1})$, it holds that $y_i\ge\tau_i$, and therefore,
\[
\mathcal{C}_{\mathrm{exact}}(\mathbf{x}_{n+1})\subseteq \bigcap_{i=1}^D \{\mathbf{y}\in\Delta^D:\ y_i\ge\tau_i\}=\mathcal{T}(\boldsymbol{\tau}).
\]

Assuming Slater’s condition holds, the KKT conditions are necessary and sufficient. The Lagrangian is
\[
\mathcal{L}(\mathbf{y},\rho,\theta,\delta)=
y_i+\rho\left(\sum_{j=1}^D y_j-1\right)
+\theta\left(t_{1-\alpha}-\sum_{j=1}^D w_j\log y_j\right)
-\sum_{j=1}^D \delta_j y_j,
\]

Solving this yields
\[
y_j=
\begin{cases}
\dfrac{\theta\,w_j}{\rho}, & j\neq i,\\[8pt]
\dfrac{\theta\,w_i}{1+\rho}, & j=i,
\end{cases}
\quad \text{and} \quad
\theta = \frac{1}{\tfrac{w_i}{1+\rho} + \tfrac{W-w_i}{\rho}}.
\]

The root $\rho$ is found from the 1D equation:
\[
F_i(\rho) = w_i \log\rho + (W-w_i)\log(1+\rho)
- W \log(w_i\rho + (W-w_i)(1+\rho)) 
+ \sum_{j=1}^D w_j \log w_j
= t_{1-\alpha}.
\]

Then,
\[
\tau_i = \frac{\theta w_i}{1+\rho}, \quad
\mathcal{T}(\mathbf{x}_{n+1}) = \{\, \mathbf{y}\in\Delta^D:\ y_i \ge \tau_i,\ i=1,\dots,D \}.
\]

\begin{figure}[ht]
    \centering
    \includegraphics[scale=0.5]{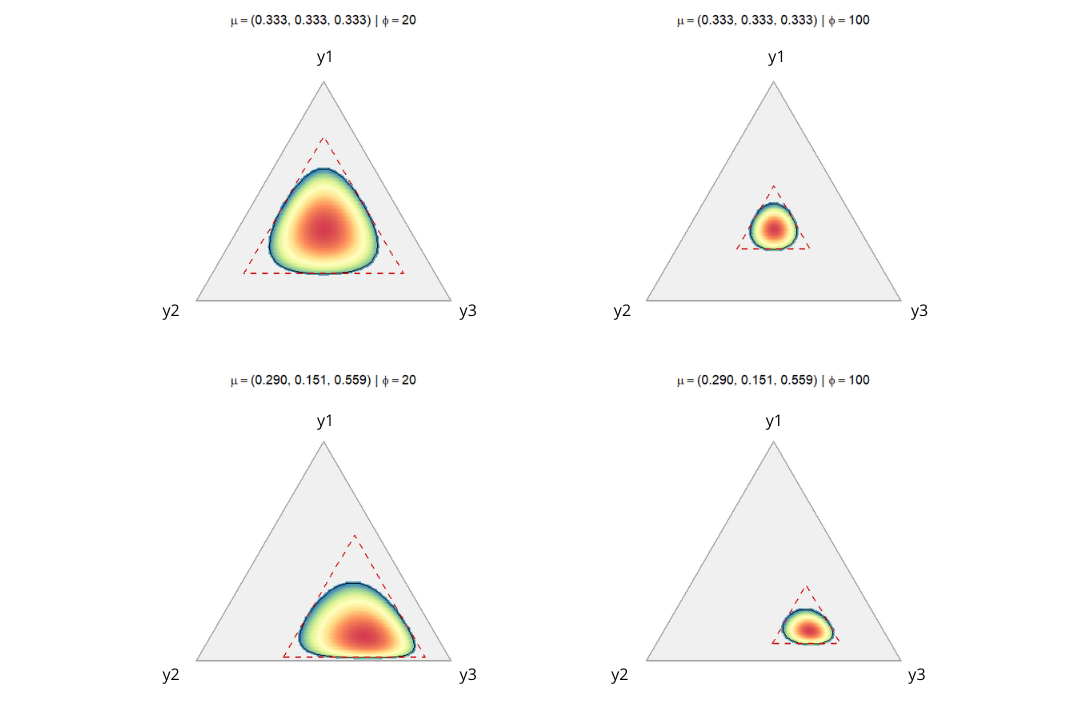}
    \caption{HDR vs. approximated triangle (red) plots for different parameter values. HDR is illustrated by color gradient within the dashed red triangle.}
    \label{hdr-aproximacao}
\end{figure}

\subsection{Refined HDR approximation via grid}

To mitigate overcoverage and avoid full-grid search over the simplex, we restrict discretization to the polytope found. Given $\boldsymbol{\tau}$, define a grid $\mathcal{Y}^* \subset \mathcal{T}(\boldsymbol{\tau})$ with step $\delta > 0$. For $k=1,\ldots,D-1$:
\[
y_k \in \{\tau_k + m\delta:\ m=0,1,2,\ldots\} \cap
\left[\tau_k,\ 1-\sum_{j=k+1}^D\tau_j-\sum_{j=1}^{k-1}y_j\right],
\]
and
\[
y_D = 1-\sum_{j=1}^{D-1}y_j.
\]
Only points with $y_D\ge\tau_D$ are kept, ensuring $\mathbf{y}\in\mathcal{T}(\boldsymbol{\tau})\subset\Delta^D$. The grid-based prediction set is:
\[
\mathcal{C}_{\textit{grid}}(\mathbf{x}_{n+1})
=
\left\{\mathbf{y}\in\mathcal{Y}^*:\ g(\mathbf{y})\ge t_{1-\alpha}\right\}.
\]

\section{Simulation Studies}
\label{sec:simulation-studies}

To evaluate the performance of the proposed method, we conducted a Monte Carlo simulation study with 1000 iterations. In each scenario, we generated samples of size $n = 1000$, where we split $70\%$ for training, $30\%$ for calibration, and used one new point for testing. In each scenario, the prediction set coverage was assessed based on a nominal coverage level ($1 - \alpha$) of $90\%$. For the grid-based discretization, we used uniformly spaced points for each component with a step size of $\delta = 0.005$.

Furthermore, we considered 3 response components in the simulations. To compute areas relative to the total simplex, we used a Monte Carlo approximation by generating 20,000 points uniformly within the full simplex using a Dirichlet distribution with all parameters equal to 1. We then checked, based on the conformal quantile $q_{1-\alpha}$, which points were accepted. The area was then estimated as the ratio of accepted points to total tested points. For the floor-polytope case, the area can be easily obtained from the equilateral triangle.

In each scenario, we generated two covariates, $x_{i22} = x_{i32} = d_{i2}$, $x_{i23} = x_{i33} = d_{i3}$, and $x_{i21} = x_{i31} = d_{i1} = 1$ for all $i$. All simulations were carried out using the R programming language and the \texttt{DirichletReg} package. The first five scenarios were considered by \cite{pereira2024class}. In scenario~1a, the covariates were drawn from a standard uniform distribution and the parameter values were chosen so that the means of $\mu_{i1}$, $\mu_{i2}$, and $\mu_{i3}$ across the $n$ observations are equal, with high response variance ($\phi$ close to 20). In this scenario, $\beta_{21} = -0.3$, $\beta_{22} = 1.0$, $\beta_{23} = -0.5$, $\beta_{31} = -0.3$, $\beta_{32} = -0.5$, $\beta_{33} = 1.0$, $\gamma_1 = 3.0$, and $\gamma_2 = \gamma_3 = 0$.

In scenarios~2a and~3a, we modified some parameter values so that the response component means become increasingly separated. In scenario~4a, we allowed $\phi$ to depend on the covariates by setting $\gamma_2 = 0.5$ and $\gamma_3 = -0.5$. In the following scenario, the first and second covariates were generated from a Bernoulli distribution with parameter $0.5$ and a Gamma distribution with parameters $3$ and $6$ (matching the mean and variance of the standard uniform distribution), respectively. Scenarios~1b to~5b mirror scenarios~1a to~5a, respectively, but with $\gamma_1 = 4.6$ ($\phi$ close to 100, corresponding to low response variance).

Additionally, to assess the effect of the number of predictors, we included scenarios~6a and~6b, where six covariates were generated from a standard uniform distribution. The regression coefficients were
\[
(\beta_{21},\beta_{22},\ldots,\beta_{27})=(0,\ 0.20,\ -0.20,\ 0.15,\ -0.15,\ 0.10,\ -0.10)^\top,
\]
\[
(\beta_{31},\beta_{32},\ldots,\beta_{37})=(0,\ -0.20,\ 0.20,\ -0.15,\ 0.15,\ -0.10,\ 0.10)^\top,
\]
that is, $\beta_3=-\beta_2$. We also adopted constant precision as in previous scenarios, with $\gamma_1 = 3.0$ and $\gamma_1 = 4.6$.

We also evaluated the robustness of the methods when the fitted model is Dirichlet but the data are generated by mechanisms that do not strictly follow a Dirichlet distribution. In each iteration, we generated $p\in\{2,6\}$ independent covariates from $U(0,1)$. Two types of misspecification were considered, each evaluated with $p=2$ and $p=6$, and under two variability levels, totaling 10 high-variability and 10 low-variability scenarios.

For scenarios~7a to~8b, the response was generated as a mixture of two Dirichlet distributions, denoted by MixDir, with probability $0.6$ for component~1 and $0.4$ for component~2. Both components share the same precision $\phi$, but have different means, constructed through two linear predictors (one for the second component and one for the third). The coefficients of component~2 are simply the negatives of those of component~1.

For $p=2$, the coefficients of component~1 are
\begin{align*}
\beta_{21}^{(1)} &= 0, & \beta_{22}^{(1)} &= 1.20, & \beta_{23}^{(1)} &= -1.20, \\
\beta_{31}^{(1)} &= 0, & \beta_{32}^{(1)} &= -1.20, & \beta_{33}^{(1)} &= 1.20,
\end{align*}
and for component~2,
\begin{align*}
\beta_{21}^{(2)} &= 0, & \beta_{22}^{(2)} &= -1.20, & \beta_{23}^{(2)} &= 1.20, \\
\beta_{31}^{(2)} &= 0, & \beta_{32}^{(2)} &= 1.20, & \beta_{33}^{(2)} &= -1.20.
\end{align*}

For $p=6$, the coefficients of component~1 are
\begin{align*}
\beta_{21}^{(1)} &= 0, & \beta_{22}^{(1)} &= 0.60, & \beta_{23}^{(1)} &= -0.60, \\
\beta_{24}^{(1)} &= 0.60, & \beta_{25}^{(1)} &= -0.60, & \beta_{26}^{(1)} &= 0.60, & \beta_{27}^{(1)} &= -0.60, \\
\beta_{31}^{(1)} &= 0, & \beta_{32}^{(1)} &= -0.60, & \beta_{33}^{(1)} &= 0.60, \\
\beta_{34}^{(1)} &= -0.60, & \beta_{35}^{(1)} &= 0.60, & \beta_{36}^{(1)} &= -0.60, & \beta_{37}^{(1)} &= 0.60.
\end{align*}
For component~2, all coefficients above are multiplied by $-1$. Precision was kept constant at two levels, with $\gamma_1\in\{3.0,\ 4.6\}$ and the remaining $\bm{\gamma}$ parameters equal to zero.

In the second type of misspecification, the compositional response is generated by a logistic-normal mechanism, denoted by LogN. Specifically, two continuous variables are simulated from a bivariate normal distribution with fixed correlation $\rho = 0.30$, and the values are mapped into the simplex using the inverse additive log-ratio (ALR) transformation. The result is a composition with three positive components summing to one. Moreover, using a positive $\rho$ induces positive correlation, which contrasts with the covariance structure of the Dirichlet distribution, thereby introducing correlation misspecification.

We considered two values for the number of covariates, $p\in\{2,6\}$. For $p=2$, the coefficients are
\[
\beta_{21}=0,\ \beta_{22}=0.90,\ \beta_{23}=-0.90,\ 
\beta_{31}=0,\ \beta_{32}=-0.90,\ \beta_{33}=0.90.
\]
For $p=6$, we used
\[
\beta_{21}=0,\ \beta_{22}=0.45,\ \beta_{23}=-0.45,\ 
\beta_{24}=0.45,\ \beta_{25}=-0.45,\ \beta_{26}=0.45,\ \beta_{27}=-0.45,
\]
\[
\beta_{31}=0,\ \beta_{32}=-0.45,\ \beta_{33}=0.45,\ 
\beta_{34}=-0.45,\ \beta_{35}=0.45,\ \beta_{36}=-0.45,\ \beta_{37}=0.45.
\]

Dispersion in the LogN mechanism is controlled by a parameter $\sigma$. In the code, $\sigma$ is numerically calibrated, separately for $p=2$ and $p=6$, to produce two variability levels comparable to a reference Dirichlet model with $\phi \approx 20$ (high variability) and $\phi \approx 100$ (low variability). A summary of the simulation scenarios is presented in Table~\ref{tabela-resumo}.

\begin{table}[ht]
\centering
\caption{Summary of simulation scenarios, correctly specified and misspecified.}
\resizebox{\textwidth}{!}{%
\begin{tabular}{lccccc}
\toprule
\textbf{Scenario} &
\textbf{Generating mechanism} &
\textbf{Mean of $(\mu_1,\mu_2,\mu_3)$} &
\textbf{Covariates} &
\textbf{$p$} &
\textbf{$\phi$} \\
\midrule
1a, 1b & Dirichlet & 0.333 \quad 0.333 \quad 0.333 & Uniform & 2 & Constant \\
2a, 2b & Dirichlet & 0.290 \quad 0.151 \quad 0.559 & Uniform & 2 & Constant \\
3a, 3b & Dirichlet & 0.308 \quad 0.049 \quad 0.643 & Uniform & 2 & Constant \\
4a, 4b & Dirichlet & 0.333 \quad 0.333 \quad 0.333 & Uniform & 2 & Variable \\
5a, 5b & Dirichlet & 0.333 \quad 0.333 \quad 0.333 & Bernoulli/Gamma & 2 & Constant \\
6a, 6b & Dirichlet & 0.333 \quad 0.333 \quad 0.333 & Uniform & 6 & Constant \\
\midrule
7a, 7b & MixDir & 0.333 \quad 0.333 \quad 0.333 & Uniform & 2 & Constant \\
8a, 8b & MixDir & 0.333 \quad 0.333 \quad 0.333 & Uniform & 6 & Constant \\
9a, 9b & LogN & 0.333 \quad 0.333 \quad 0.333 & Uniform & 2 & $(\sigma)$ Constant \\
10a, 10b & LogN & 0.333 \quad 0.333 \quad 0.333 & Uniform & 6 & $(\sigma)$ Constant \\
\bottomrule
\end{tabular}
}
\label{tabela-resumo}
\end{table}

We also conducted two additional experiments to evaluate the computational efficiency of the HDR-approximation method with grid search relative to a naive grid search over the entire simplex. Two different step sizes were used: $\delta = 0.02$ for the HDR-approximation method and $\delta = 0.005$ for the naive approach.

In the first scenario of this experiment, we adopted the same structure as in scenarios~1a and~1b. However, as the number of response components increases, the computational cost of grid search is expected to increase as well. Therefore, we considered a case with four response components. Specifically,
\begin{align*}
    \beta_{21} &= -0.3, & \beta_{22} &= 1.0,  & \beta_{23} &= -0.5, \\
    \beta_{31} &= -0.3, & \beta_{32} &= -0.5, & \beta_{33} &= 1.0,  \\
    \beta_{41} &= -0.3, & \beta_{42} &= 0.5,  & \beta_{43} &= 0.5,  \\
    \gamma_1   &= 3.0,  & \gamma_2   &= \gamma_3 = \gamma_4 = 0.
\end{align*}

This parameter configuration yields component means equal to $0.2335$, $0.1816$, $0.2850$, and $0.2999$. The two covariates were generated from a standard uniform distribution, similarly to the previous scenarios. Since these are correctly specified scenarios, the results will be reported in the corresponding subsection.

\subsection{Results: Correct Specification}
Table \ref{tab:results_correct_spec} presents the simulation results for the correctly specified scenarios, where the data-generating mechanism coincides with the Dirichlet model used for fitting. In all scenarios, the nominal coverage level was set at 90\%. Overall, the three evaluated procedures effectively control the empirical coverage, although they exhibit systematic differences regarding the degree of conservatism, predictive set size, and computational cost.

The quantile residuals method presents empirical coverage close to the nominal level in most scenarios, with values typically ranging between 89\% and 92\%. Small oscillations around 90\% are observed, especially in configurations with greater asymmetry among the response components, but without indicating a systematic loss of coverage control. This behavior is consistent with the method's construction, which combines marginal intervals to form the joint predictive set.

In contrast, the HDR-aprox method displays a clearly more conservative pattern. In all correctly specified scenarios, the empirical coverage exceeds the nominal level, varying approximately between 93.5\% and 96.1\%. This increase in coverage stems from approximating the highest density region using a polytope defined by floors, which tends to include additional regions of the simplex. Consequently, the predictive sets generated by this method present substantially larger relative areas, particularly in scenarios where the means of the response components are further apart.

The HDR-aprox-grid method yielded coverages between 89.3\% and 91.4\%, achieving the specified 90\% nominal level. Even in scenarios of higher variability and asymmetry, the method demonstrated its ability to maintain coverage. Thus, it is evident that this method corrects the over-coverage inherent in the HDR-aprox method, which is essential for obtaining more efficient regions.

Regarding the relative area of the prediction sets, a clear pattern emerges among the methods. HDR-aprox produces the largest sets, reflecting its conservative nature. The quantile residuals method generates considerably smaller sets, yet consistently with slightly larger relative areas than those obtained by HDR-aprox-grid. In turn, HDR-aprox-grid presents the smallest relative areas among the considered methods across all analyzed scenarios.

Furthermore, the expected effect of response variability is verified: in scenarios with lower dispersion (scenarios “b”, associated with higher $\phi$ values), the relative areas consistently decrease for all methods, indicating a higher concentration of probability mass within the simplex.

Regarding computational cost, the quantile residuals and HDR-aprox methods show low and comparable average execution times, generally under two hundredths of a second per repetition, with the latter being slightly faster. Thus, they are fully suitable for applications with a large number of iterations. In contrast, the HDR-aprox-grid method presents a higher computational cost, with average execution times about 10 times higher than those observed for the quantile residual method. Nevertheless, the observed times remain moderate within the context of the simulations performed.

In summary, under correct model specification, the results indicate a clear trade-off between methods. The quantile residuals procedure offers good coverage with relatively compact sets and low computational cost, although slightly larger than those obtained by HDR-aprox-grid. HDR-aprox guarantees high coverage, but at the cost of wide predictive sets. Finally, HDR-aprox-grid emerges as an intermediate alternative, capable of combining coverage close to the nominal level with more geometrically efficient predictive sets, albeit with a higher computational cost.

\begin{table}[!htb]
\centering
\caption{Simulation results for correctly specified scenarios. Metrics reported are empirical coverage, mean relative area, and execution time.}
\label{tab:results_correct_spec}
\begin{tabular}{llccc}
\toprule
\textbf{Scenario} & \textbf{Method} & \textbf{\begin{tabular}[c]{@{}c@{}}Empirical \\ Coverage (\%)\end{tabular}} & \textbf{\begin{tabular}[c]{@{}c@{}}Relative\\ Area\end{tabular}} & \textbf{\begin{tabular}[c]{@{}c@{}}Avg. Time\\ (s)\end{tabular}} \\ \midrule

\multirow{3}{*}{\textbf{1a}} & Quantile Res. & 90.1 & 0.2433 & 0.0139 \\
                             & HDR-aprox        & 94.7 & 0.3221 & 0.0106 \\
                             & HDR-aprox-grid  & 90.3 & 0.2339 & 0.1532 \\ \midrule

\multirow{3}{*}{\textbf{1b}} & Quantile Res. & 89.5 & 0.0537 & 0.0175 \\
                             & HDR-aprox        & 94.5 & 0.0802 & 0.0101 \\
                             & HDR-aprox-grid  & 90.1 & 0.0509 & 0.1753 \\ \midrule

\multirow{3}{*}{\textbf{2a}} & Quantile Res. & 92.1 & 0.1985 & 0.0206 \\
                             & HDR-aprox        & 94.4 & 0.2709 & 0.0123 \\
                             & HDR-aprox-grid  & 91.4 & 0.1796 & 0.1923 \\ \midrule

\multirow{3}{*}{\textbf{2b}} & Quantile Res. & 89.9 & 0.0437 & 0.0169 \\
                             & HDR-aprox        & 94.6 & 0.0674 & 0.0106 \\
                             & HDR-aprox-grid  & 89.3 & 0.0405 & 0.1686 \\ \midrule

\multirow{3}{*}{\textbf{3a}} & Quantile Res. & 88.8 & 0.0962 & 0.0144 \\
                             & HDR-aprox        & 96.1 & 0.6910 & 0.0081 \\
                             & HDR-aprox-grid  & 89.4 & 0.0539 & 0.1387 \\ \midrule

\multirow{3}{*}{\textbf{3b}} & Quantile Res. & 91.1 & 0.0179 & 0.0143 \\
                             & HDR-aprox        & 95.4 & 0.2075 & 0.0084 \\
                             & HDR-aprox-grid  & 90.7 & 0.0137 & 0.1381 \\ \midrule

\multirow{3}{*}{\textbf{4a}} & Quantile Res. & 90.1 & 0.1924 & 0.0145 \\
                             & HDR-aprox        & 95.5 & 0.3195 & 0.0085 \\
                             & HDR-aprox-grid  & 89.8 & 0.1718 & 0.1388 \\ \midrule

\multirow{3}{*}{\textbf{4b}} & Quantile Res. & 89.4 & 0.0429 & 0.0143 \\
                             & HDR-aprox        & 93.5 & 0.0667 & 0.0077 \\
                             & HDR-aprox-grid  & 89.4 & 0.0398 & 0.1390 \\ \midrule

\multirow{3}{*}{\textbf{5a}} & Quantile Res. & 89.3 & 0.2320 & 0.0144 \\
                             & HDR-aprox        & 94.6 & 0.3083 & 0.0093 \\
                             & HDR-aprox-grid  & 89.6 & 0.2210 & 0.1425 \\ \midrule

\multirow{3}{*}{\textbf{5b}} & Quantile Res. & 90.1 & 0.0511 & 0.0150 \\
                             & HDR-aprox        & 94.6 & 0.0770 & 0.0093 \\
                             & HDR-aprox-grid  & 90.2 & 0.0483 & 0.1397 \\ \midrule

\multirow{3}{*}{\textbf{6a}} & Quantile Res. & 90.5 & 0.2537 & 0.0177 \\
                             & HDR-aprox        & 95.0 & 0.3361 & 0.0104 \\
                             & HDR-aprox-grid  & 90.1 & 0.2468 & 0.1891 \\ \midrule

\multirow{3}{*}{\textbf{6b}} & Quantile Res. & 90.1 & 0.0563 & 0.0164 \\
                             & HDR-aprox        & 96.0 & 0.0838 & 0.0103 \\
                             & HDR-aprox-grid  & 90.5 & 0.0538 & 0.1703 \\ \bottomrule
\end{tabular}
\end{table}

Finally, Table \ref{tab:hdr-vs-simplex} presents comparative results between the HDR-aprox method with approximation and grid discretization and the more naive approach based on exhaustive search over the simplex (Simplex-grid), used as a reference. For discretization within the polytope, we used $\delta \approx 0.02$, and for the naive method, $\delta \approx 0.005$.

The data shows that, even with a modest grid, the HDR-aprox-grid method produced smaller relative areas and volumes in all evaluated scenarios. In the case with three components, the average execution time was less than 25\% of that required for the Simplex-grid, demonstrating a significant advantage in terms of computational efficiency.

Empirical coverage, while important in other comparisons, is not the central focus here. Due to the high cost of the simplex search with 4 components and 1000 iterations, we chose to reduce this number to 100 iterations only for this specific experiment.

For four components, the computational cost of the Simplex-grid grows sharply, while the HDR-aprox-grid, although also showing an increase in time, remains much more efficient. The computational cost of the Simplex-grid in these cases is more than 50 times that of the HDR-aprox-grid. Additionally, the relative areas and volumes are smaller for the HDR-aprox-grid. These results justify the preference for HDR-aprox-grid over Simplex-grid.

\begin{table}[!htb]
\centering
\caption{Performance comparison: HDR-aprox-grid vs. Simplex-grid (3 and 4 components).}
\label{tab:hdr-vs-simplex}
\small
\setlength{\tabcolsep}{8pt}
\begin{tabular}{ll c c c}
\toprule
\multirow{2}{*}{\textbf{$\phi$}} & \multirow{2}{*}{\textbf{Method}} & \textbf{Cov.} & \textbf{Rel. Area/} & \textbf{Avg. Time} \\
 & & \textbf{(\%)} & \textbf{Volume} & \textbf{(s)} \\
\midrule

\multicolumn{5}{l}{\textbf{Scenario: 3 components}} \\
\addlinespace[2pt]
$\approx 20$  & HDR-aprox-grid & 89.0 & 0.2160 & 0.0240 \\
              & Simplex-grid   & 89.0 & 0.2330 & 0.1100 \\
\addlinespace

$\approx 100$ & HDR-aprox-grid & 90.0 & 0.0433 & 0.0160 \\
              & Simplex-grid   & 90.0 & 0.0514 & 0.0993 \\
\midrule

\multicolumn{5}{l}{\textbf{Scenario: 4 components}} \\
\addlinespace[2pt]
$\approx 20$  & HDR-aprox-grid & 87.0 & 0.1770 & 0.2930 \\
              & Simplex-grid   & 87.0 & 0.1970 & 15.5000 \\
\addlinespace

$\approx 100$ & HDR-aprox-grid & 95.0 & 0.0173 & 0.1050 \\
              & Simplex-grid   & 95.0 & 0.0216 & 16.4000 \\
\bottomrule
\end{tabular}
\end{table}

\subsection{Results: Misspecification}
Table \ref{tab:tabela-ma-especif} presents the simulation results for the misspecified scenarios, where the fitted model assumes a Dirichlet distribution, but the data-generating mechanism does not follow this distribution exactly, covering both mixtures of Dirichlet distributions and a mechanism based on the logistic normal. As in the correctly specified scenarios, the nominal coverage level considered was 90\%.

In general, it is observed that the methods maintain satisfactory performance in terms of empirical coverage even under model misspecification, highlighting the robustness of the conformal procedures considered. The quantile residuals method presents empirical coverages very close to the nominal level in all analyzed scenarios, typically ranging between 89\% and 92\%. This behavior suggests that even when the assumed distribution in the fit is incorrect, the method manages to preserve coverage control, albeit based on marginal constraints.

The HDR-aprox method continues to present a more conservative pattern under misspecification. In all considered scenarios, empirical coverages exceed the nominal level, reaching values between approximately 93\% and 96\%. This result indicates that constructing the predictive set from an approximation of the highest density region maintains an additional safety margin even when the functional form of the density is incorrect. However, this conservatism is again accompanied by wider predictive sets.

Regarding the HDR-aprox-grid method, it is observed that, as in the correctly specified scenarios, HDR-aprox-grid is solid in terms of coverage, ranging between 89.5\% and 91.3\%, reducing the over-coverage of the original method and meeting the specified nominal coverage.

This behavior is clearly reflected in the mean relative areas. Compared to the correctly specified scenarios, the relative areas are substantially larger, especially in scenarios based on the Dirichlet mixture (7a to 8b), which was expected given the increased complexity of the generating mechanism. The HDR-aprox method systematically produces the largest relative areas, reflecting the inclusion of additional regions of the simplex due to the floor approximation in a misspecification context.

Furthermore, in general, HDR-aprox-grid produces smaller relative areas than the quantile residuals method, but for the Dirichlet mixture, this reverses in cases 7b and 8b, unlike what was observed in the correctly specified scenarios, where HDR-aprox-grid achieved nominal coverage and produced the smallest relative areas for all evaluated situations.

As in the correctly specified scenarios, the effect of response variability is also observed. In scenarios with lower dispersion (“b” scenarios), the relative areas are consistently smaller for all methods, while empirical coverages remain stable, indicating that reduced variability concentrates the probability mass and results in more compact predictive sets, even under misspecification.

Regarding computational cost, the pattern is similar to previous scenarios. The quantile residuals and HDR-aprox methods present low and comparable average execution times, while HDR-aprox-grid incurs a substantially higher computational cost due to the additional grid search. Nonetheless, average times remain moderate and compatible with simulation studies and practical applications.

In summary, the results under misspecification indicate that the evaluated methods are robust in maintaining empirical coverage close to or above the nominal level. The quantile residuals method offers good stability and low computational cost, HDR-aprox guarantees high coverage at the expense of wider sets, and HDR-aprox-grid again stands out for producing more efficient predictive sets in terms of relative area while maintaining adequate coverage even when the fitted model does not coincide with the data-generating mechanism.

\begin{table}[!htb]
\centering
\caption{Simulation results for misspecified scenarios. Metrics reported are empirical coverage, mean relative area, and average execution time for each method.}
\begin{tabular}{llccc}
\toprule
\textbf{Scenario} & \textbf{Method} & \textbf{\begin{tabular}[c]{@{}c@{}}Empirical \\ Coverage (\%)\end{tabular}} & \textbf{\begin{tabular}[c]{@{}c@{}}Relative\\ Area\end{tabular}} & \textbf{\begin{tabular}[c]{@{}c@{}}Avg. Time\\ (s)\end{tabular}} \\ \midrule

\multirow{3}{*}{\textbf{7a}} & Quantile Res. & 90.4 & 0.5664 & 0.0172 \\
                             & HDR-aprox        & 93.0 & 0.6696 & 0.0119 \\
                             & HDR-aprox-grid  & 89.5 & 0.5583 & 0.1896 \\ \midrule

\multirow{3}{*}{\textbf{7b}} & Quantile Res. & 90.8 & 0.4068 & 0.0210 \\
                             & HDR-aprox        & 94.8 & 0.5281 & 0.0131 \\
                             & HDR-aprox-grid  & 91.1 & 0.4177 & 0.2263 \\ \midrule

\multirow{3}{*}{\textbf{8a}} & Quantile Res. & 89.6 & 0.4987 & 0.0195 \\
                             & HDR-aprox        & 93.8 & 0.6029 & 0.0141 \\
                             & HDR-aprox-grid  & 90.6 & 0.4902 & 0.2139 \\ \midrule

\multirow{3}{*}{\textbf{8b}} & Quantile Res. & 90.6 & 0.3336 & 0.0195 \\
                             & HDR-aprox        & 94.0 & 0.4436 & 0.0123 \\
                             & HDR-aprox-grid  & 90.3 & 0.3399 & 0.2279 \\ \midrule

\multirow{3}{*}{\textbf{9a}} & Quantile Res. & 91.6 & 0.2489 & 0.0183 \\
                             & HDR-aprox        & 96.3 & 0.3264 & 0.0102 \\
                             & HDR-aprox-grid  & 91.3 & 0.2379 & 0.1729 \\ \midrule

\multirow{3}{*}{\textbf{9b}} & Quantile Res. & 91.6 & 0.0567 & 0.0171 \\
                             & HDR-aprox        & 96.1 & 0.0841 & 0.0112 \\
                             & HDR-aprox-grid  & 90.9 & 0.0535 & 0.1699 \\ \midrule

\multirow{3}{*}{\textbf{10a}} & Quantile Res. & 90.2 & 0.2551 & 0.0221 \\
                              & HDR-aprox        & 95.3 & 0.3345 & 0.0147 \\
                              & HDR-aprox-grid  & 89.8 & 0.2448 & 0.2424 \\ \midrule

\multirow{3}{*}{\textbf{10b}} & Quantile Res. & 90.0 & 0.0548 & 0.0194 \\
                              & HDR-aprox        & 95.8 & 0.0816 & 0.0149 \\
                              & HDR-aprox-grid  & 90.2 & 0.0519 & 0.2109 \\ \bottomrule
\end{tabular}
\label{tab:tabela-ma-especif}
\end{table}

\section{Application}
\label{sec:application}
To assess the performance of the methods on real datasets, we consider two distinct applications. Section \ref{sec:sono} presents an analysis applied to sleep-stage data, while Section \ref{sec:biomass} addresses the problem of biomass allocation. As in the simulations, we consider two scenarios: one with correct model specification and another with model misspecification.

In both cases, we conduct an initial exploratory analysis, followed by model fitting and diagnostic assessment via simulated envelopes, based on the residual approach proposed by \cite{pereira2024class}.

The main goal of the applications is to carry out a parametric regression analysis combined with a predictive task, focusing on evaluating the methods’ performance in real-world contexts. Although a traditional inferential analysis is also included, the focus is not on achieving the best possible fit, but rather on investigating the behavior of the proposed methods in practical scenarios. 

\subsection{Application 1: sleep stages}
\label{sec:sono}

The distribution of total sleep time across stages N1, N2, N3, and REM can be understood as a composition, since the parts must sum to 100\%. Functionally, particular attention is often given to stages N3 and REM because they are associated, respectively, with physiological recovery and important psychological processes; therefore, they are expected to occupy a relevant fraction of total sleep time.

From a physiological standpoint, N1 and N2 are typically classified as light sleep \citep{ritmala2015sleep}. N3 corresponds to deep sleep, the phase in which greater restoration of muscles and tissues occurs \citep{hussain2022quantitative}. REM is recognized by rapid eye movements and autonomic changes, such as faster breathing, and it plays a prominent role in aspects related to emotional processing \cite{tempesta2018sleep}. As a reference, \cite{gonzalez2019analysis} report intervals considered usual for the relative time in each stage: 3--8\% in N1, 45--55\% in N2, 15--20\% in N3, and 20--25\% in REM.

Despite this compositional nature, it is common for studies to treat stage proportions as if they were independent, conducting separate analyses for each of them \cite{gonzalez2019analysis,maski2021stability}. This approach can ignore the fact that increasing the share of one stage necessarily reduces that of at least one other stage, since the total is fixed.

In this application, we use the data from \cite{veje2021sleep}, who evaluated 42 adults---22 with tick-borne encephalitis (TBE) and 20 healthy controls---with the aim of investigating possible impacts of the disease on sleep quality. Here, we adopt a Dirichlet regression model to examine whether the composition of sleep time across stages changes as a function of total sleep time (TST) and the presence of TBE.

Because this work also aims to evaluate the proposed methods in realistic scenarios, we sought to apply SCP to a real dataset. However, the literature does not provide a dataset with a sufficiently large sample size that is well fitted by Dirichlet regression.

Therefore, we opted to expand the sleep-stage dataset. First, we fit the correctly specified model using the 42 original observations and using TST as the covariate (since the covariate indicating the presence of TBE is not significant). Next, we performed a \textit{bootstrap} resampling of the covariate. For each resampled covariate value, we associate an estimated mean and precision obtained from the fitted model.

To form a final dataset with 1000 observations, we used the 42 original observations and generated the remaining 958 as follows: we sampled a covariate value with replacement, used the corresponding estimated parameters, and generated a new synthetic response from them. This procedure was repeated until reaching the desired total of 1000 observations.

\subsection{Exploratory analysis}
Table \ref{tab:descritiva_sono} presents descriptive statistics for the proportions of time spent in each sleep stage (N1, N2, N3, and REM), as well as total sleep time (TST), considering the 42 individuals in the study. On average, most time is devoted to stage N2 (mean = 0,5336), in line with values expected from the literature \cite{gonzalez2019analysis}. Stages N3 and REM, which are functionally more relevant, present means close to the lower limit of the normality intervals: 13,4\% and 19,2\%, respectively. Stage N1, in turn, represents on average 14\% of sleep time, with considerable variation (from 0,4\% to 46,6\%), which may suggest instability or sleep fragmentation in some individuals.

\begin{table}[ht]
    \centering
    \caption{Descriptive Statistics of Sleep Stages and Total Sleep Time}
    \label{tab:descritiva_sono}
    \begin{tabular}{lcccccc}
        \toprule
        \textbf{Variable} & \textbf{Min.} & \textbf{1st Q.} & \textbf{Median} & \textbf{Mean} & \textbf{3rd Q.} & \textbf{Max.} \\ 
        \midrule
        N1  & 0,0040 & 0,0788 & 0,1307 & 0,1403 & 0,1846 & 0,4663 \\
        N2  & 0,1967 & 0,4632 & 0,5335 & 0,5336 & 0,6102 & 0,8804 \\
        N3  & 0,0010 & 0,0707 & 0,1217 & 0,1341 & 0,1859 & 0,4907 \\
        REM & 0,0211 & 0,1274 & 0,1814 & 0,1920 & 0,2490 & 0,4920 \\
        \midrule
        TST & 3,877  & 5,622  & 6,613  & 6,481  & 7,217  & 8,692  \\
        \bottomrule
    \end{tabular}
\end{table}

\begin{figure}
    \centering
    \includegraphics[scale = 0.65]{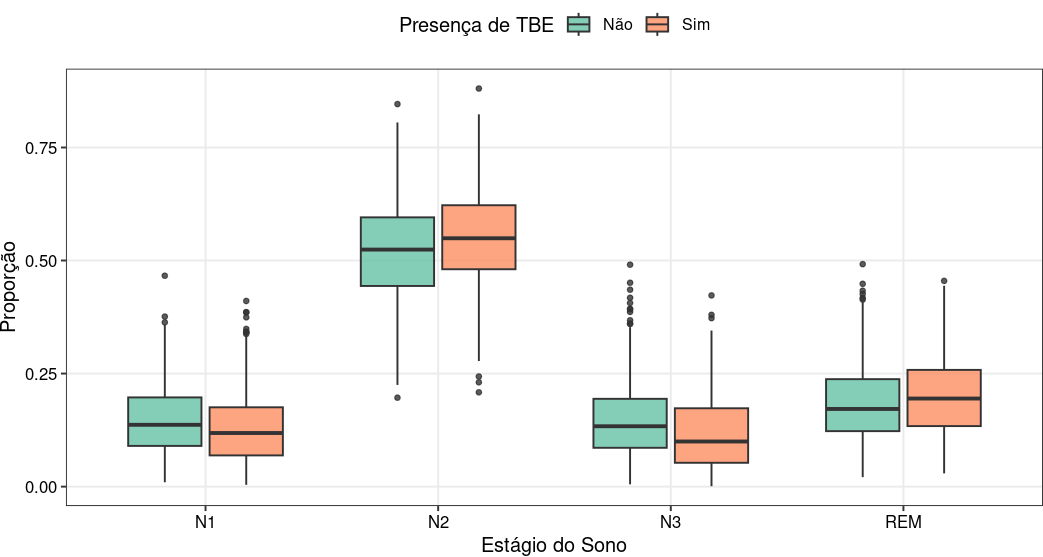}
    \caption{Distribution of sleep-stage proportions across the control and TBE groups.}
    \label{fig:boxplot-sleep}
\end{figure}
Figure \ref{fig:boxplot-sleep} shows the distribution of the proportions of each sleep stage by group. For all stages, there is little difference between the boxplots for individuals with and without TBE. Thus, it does not seem that the distribution of time proportions in each sleep stage varies as a function of the presence or absence of TBE.

\begin{figure}
    \centering
    \includegraphics[scale = 0.5]{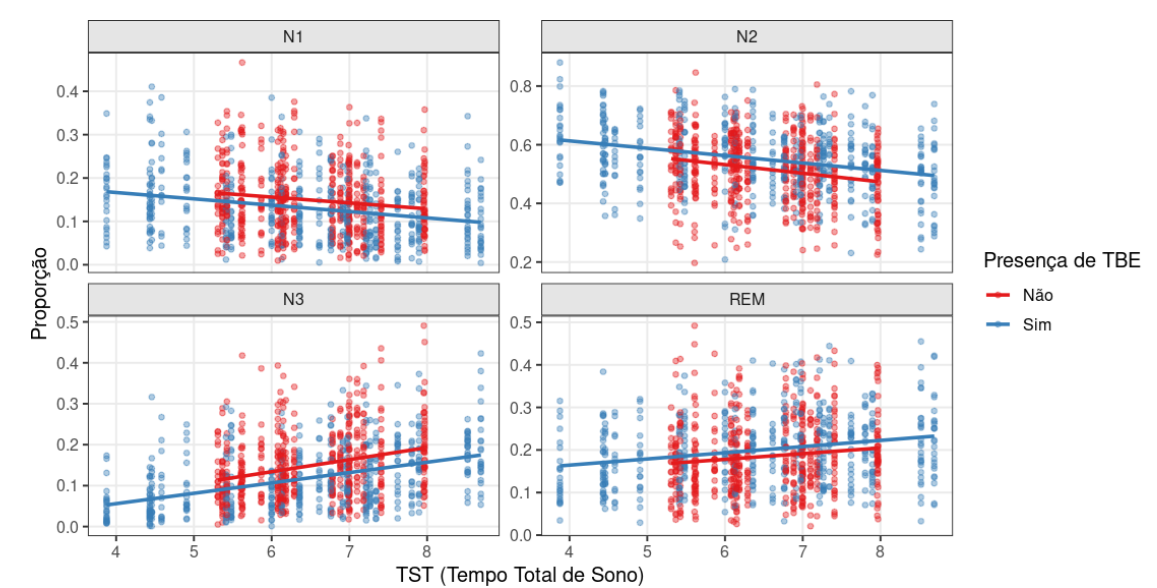}
    \caption{Relationship between total sleep time (TST) and the proportion of time in each sleep stage, by group (control and TBE).}
    \label{fig:plot-sleep-2}
\end{figure}
Figure \ref{fig:plot-sleep-2} illustrates the relationship between total sleep time (TST) and the proportion of time spent in each stage, separating the control group (0) and the TBE group (1). A general upward trend is observed for the proportions of N3 and REM as TST increases, while stages N1 and N2 tend to decrease, especially in the TBE group. This redistribution suggests that, as individuals sleep longer, there is a favorable shift toward the more restorative sleep stages, particularly in the group affected by the disease.

\subsection{Model fitting}
For model fitting we considered the final model used in \cite{pereira2024class}, in which both the mean vector of the proportions of time spent in each sleep stage and the precision were modeled by total sleep time. This is because, based on the likelihood ratio test, the authors concluded that neither healthy adults nor adults with TBE showed a significant difference in the proportion of time spent in each sleep stage.

Table~\ref{tab:coeficientes_dirichlet_sono} presents the estimated coefficients of the Dirichlet regression model fitted to the proportions of time spent in each sleep stage---N2, N3, and REM---with stage N1 taken as the reference category. The model includes total sleep time (TST) as a covariate common to all submodels, in addition to the precision submodel $(\phi)$.

For the N2 submodel, we observe a positive coefficient for TST, implying that greater total sleep time is associated with an increase in the relative proportion of N2 compared to the reference stage. It is estimated that the ratio between the mean time in stage N2 and the mean time in stage N1 increases by 12,1\% for each additional hour of TST.

In the N3 submodel, we again observe a positive coefficient for TST; it is estimated that the ratio between the mean time in stage N3 and the mean time in stage N1 increases by 31,6\% for each additional hour of TST. This result reinforces the positive association between deep sleep and total sleep duration.

For the REM stage, it is estimated that the ratio between the mean time in this stage and the mean time in stage N1 increases by 20,4\% for each additional hour of TST, indicating that longer sleep periods are associated with larger relative proportions of REM, a stage essential for emotional processing and memory consolidation.

The precision submodel $(\phi)$ yields an interesting result: the TST coefficient is negative, suggesting that as total sleep time increases there is a slight reduction in the precision of the response variable, indicating greater individual variability in sleep composition over longer sleep periods. This may reflect inter-individual differences in sleep architecture, especially in deeper stages and REM periods, as total duration increases.

Taken together, the results indicate that total sleep time plays a decisive role in the composition of sleep stages, especially in N3 and REM, which are more restorative. The use of Dirichlet regression appears appropriate to capture the proportional and interdependent nature of these data, revealing how changes in total sleep time affect the relative distribution across sleep stages.

\begin{table}[ht]
\centering
\caption{Parameter estimates and standard errors in the model for the proportion of time spent in each sleep stage.}
\begin{tabular}{ccccc}
\hline
Submodel & Covariate & Estimate & Std. Error & Exp(estim) \\
\hline
$\mu_{N2}$ & Intercept & 0,0800 & 0,2315 & 1,083 \\
           & TST       & 0,1146 & 0,0355 & 1,121 \\
\hline
$\mu_{N3}$ & Intercept & -1,7599 & 0,2487 & 0,172 \\
           & TST       & 0,2745 & 0,0379 & 1,316 \\
\hline
$\mu_{REM}$& Intercept & -1,2929 & 0,2486 & 0,274 \\
           & TST       & 0,1857 & 0,0379 & 1,204 \\
\hline
$\phi$     & Intercept & 0,8955 & 0,1294 & 2,449 \\
           & TST       & -0,0329 & 0,0196 & 0,968 \\
\hline
\end{tabular}

\label{tab:coeficientes_dirichlet_sono}
\end{table}

To complete the inferential analysis, we carried out a brief diagnostic assessment using the residual proposed and recommended by \cite{pereira2024class} for compositional data. Figure 7 presents the normal probability plot with a simulated envelope for this residual. The plot does not suggest inadequacy of the Dirichlet regression model for the proportion of time spent in each sleep stage, with total sleep time as the covariate in all submodels.

\begin{figure}
    \centering
    \includegraphics[scale=0.5]{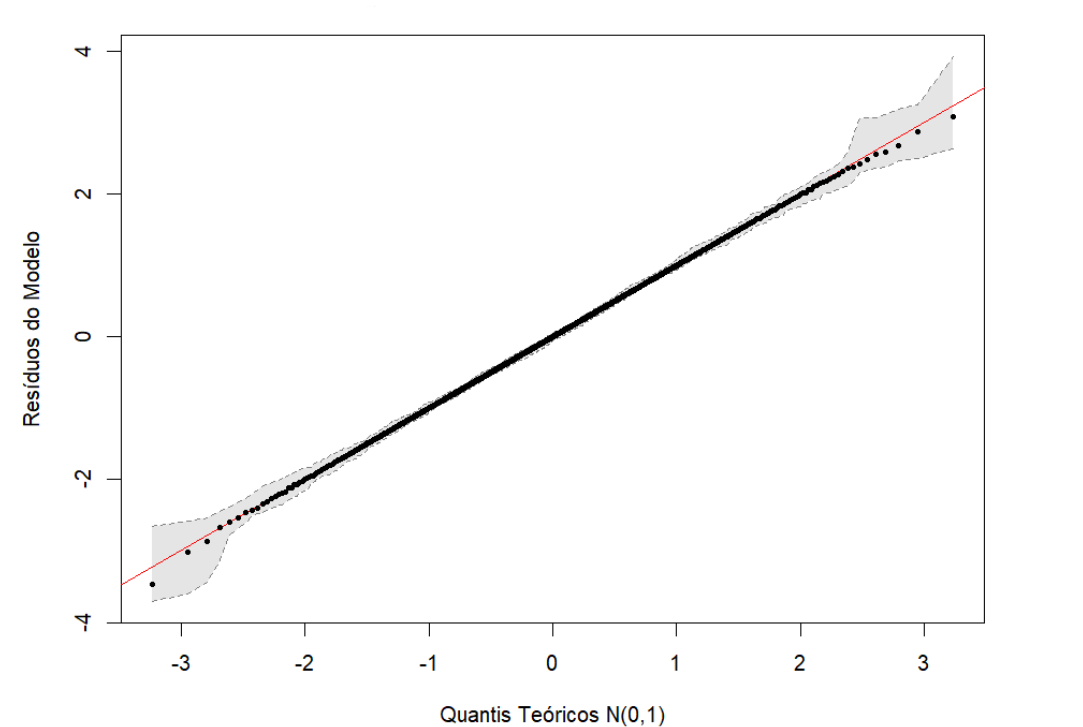}
    \caption{Normal probability plot with simulated envelope for the residuals.}
    \label{fig:qqplot-sleep}
\end{figure}

\subsection{Predictive analysis}

Table~\ref{tab:preditivos_sono_atualizada} compares three methods for constructing predictive sets---Quantile, HDR-aprox, and HDR-aprox-grid---evaluated in terms of mean relative volume and empirical coverage. We used 70\% for training, 20\% for calibration, and 10\% for testing, with a nominal coverage level of 90\%. In addition, to reduce variability in the results, we repeated this split procedure 5 times.

We observe that the quantile method achieves an empirical coverage of 89,6\%, very close to the nominal level, with an intermediate relative volume. This result indicates that, although the method controls coverage adequately, it does so through relatively wide regions, reflecting the marginal nature of the procedure, which does not fully exploit the compositional dependence structure in the data.

The HDR-aprox method, in turn, yields the highest empirical coverage among the three procedures (97,8\%), substantially exceeding the nominal level. This behavior indicates a conservative procedure in which coverage is guaranteed at the expense of more voluminous predictive regions. Indeed, the mean relative volume associated with this method is substantially larger, suggesting that the analytical approximation of the HDR tends to include additional portions of the simplex that are not strictly necessary to achieve the desired coverage.

In contrast, HDR-aprox-grid attains an empirical coverage of 90,7\%, higher than that of the quantile method, while having the smallest mean relative volume among the evaluated methods. This result highlights an important efficiency gain, indicating that combining the HDR approximation with verification on a physical grid allows one to construct more concentrated predictive regions without compromising coverage control. In practical terms, this method produces more informative predictive sets, capturing more precisely the highest-density regions of the conditional distribution on the simplex.

Overall, the results corroborate the classic \textit{trade-off} between coverage and predictive-region volume. While HDR-aprox prioritizes safety in terms of coverage, HDR-aprox-grid stands out for offering a balance between coverage and relative volume. In the sleep-stage context, where joint interpretation of proportions is fundamental, the latter method is particularly attractive because it produces more compact regions coherent with the compositional structure of the data.

\begin{table}[!htb]
\centering
\caption{Performance of predictive methods in terms of empirical coverage and mean relative volume.}
\label{tab:preditivos_sono_atualizada}
\begin{tabular}{lcc}
\hline
\textbf{Method} & \textbf{Empirical Coverage (\%)} & \textbf{Relative volume} \\ \hline
Quantile       & 89,6                             & 0,182                    \\
HDR-aprox      & 97,8                             & 0,312                    \\
HDR-aprox-grid & 90,7                             & 0,141                    \\ \hline
\end{tabular}
\end{table}

\subsection{Application 2: biomass allocation} 
\label{sec:biomass}
For this application we use the dataset presented in \cite{douma2019analysing}. In this context, plants are known to allocate biomass differently among leaves, stems, roots, and reproductive structures, following ontogenetic trajectories that interact with the prevailing climate.

Several methodological approaches are used to analyze the resulting allocation patterns, including computing proportions or fractions of biomass in different organs at a given time, as well as allometric analyses of data collected across species or over experimental growth periods.

The dataset analyzed comes from an experiment with two plant species with different growth rates: \textit{Deschampsia flexuosa} (slow-growing) and \textit{Holcus lanatus} (fast-growing). Plants were cultivated under two levels of nitrate supply (high and low) for up to 49 days. During this period, individuals were harvested at different times for measurement of biomass allocated to leaves, stems, and roots, totaling 500 observations.

The response variables correspond to the proportions of total biomass allocated to leaves (LMF), stems (SMF), and roots (RMF). These proportions were modeled simultaneously via Dirichlet regression, considering as explanatory variables: species, nitrate level, time (including a quadratic term), total biomass, and their interactions. Model precision was also fitted based on these variables.

\subsection{Exploratory analysis}

\begin{figure}[!htb]
    \centering
    \includegraphics[scale=0.5]{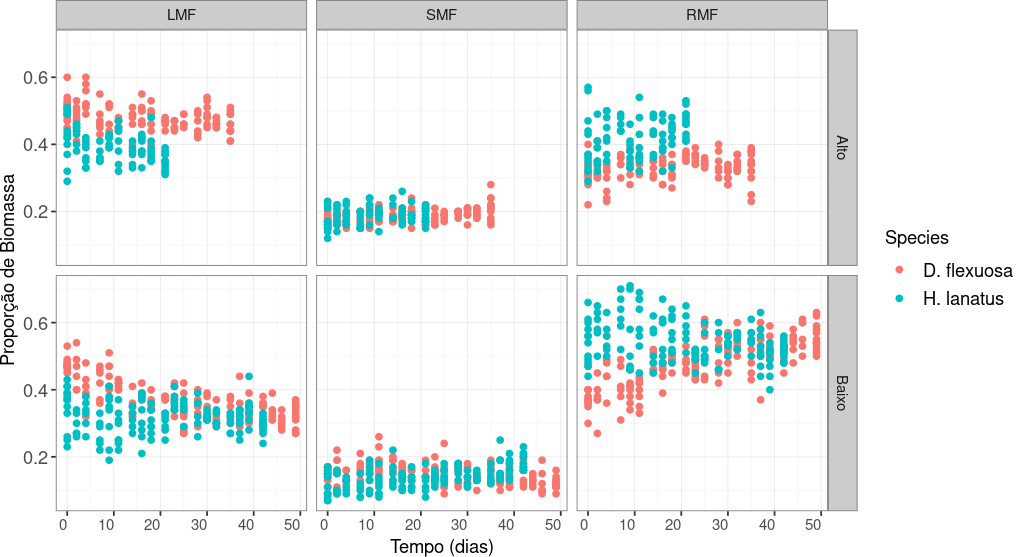}
    \caption{Proportion of biomass allocated to leaves (LMF), stems (SMF), and roots (RMF) over time for the species Deschampsia flexuosa (red) and Holcus lanatus (blue), cultivated under two nitrate-supply levels: high (top panel) and low (bottom panel).}
    \label{fig:biomassa}
\end{figure}

Figure~\ref{fig:biomassa} shows the evolution over days of the proportions of biomass allocated to leaves (LMF), stems (SMF), and roots (RMF) for the species \textit{Deschampsia flexuosa} (in red) and \textit{Holcus lanatus} (in blue), cultivated under two nitrate-supply levels: high (top panel) and low (bottom panel).

Under high nitrate, \textit{D. flexuosa} tends to maintain higher values of leaf allocation compared with \textit{H. lanatus}. Over time, both species show a slight reduction in this proportion. Regarding stem allocation (SMF), both species exhibit similar patterns, with relatively stable proportions over time and a modest increase in the final days. For root allocation (RMF), \textit{H. lanatus} shows a higher proportion of biomass in roots, and the mean of this proportion does not appear to vary much over time. In contrast, \textit{D. flexuosa} maintains lower and stable values throughout the experimental period.

Under low nitrate supply, a different behavior is observed. Leaf allocation (LMF) remains relatively constant for both species, with \textit{D. flexuosa} presenting slightly higher proportions. Stem allocation (SMF) patterns remain similar to those under high nitrate, with no pronounced variation. In contrast, root allocation (RMF) shows more marked differences between species. \textit{H. lanatus} begins the experiment with a higher proportion of biomass allocated to roots, a difference that decreases over time. Conversely, \textit{D. flexuosa} shows a gradual increase in root allocation over the 49 days. These patterns suggest that the two species respond differently to nitrogen limitation, particularly in how they distribute biomass between roots and aboveground tissues.

Overall, the results reflect a clear interaction among species, time, and nutritional condition, which reinforces the choice of models capable of accounting for these complex interactions, such as the Dirichlet regression adopted in the analysis. Differences in root allocation under low nitrate supply indicate contrasting strategies for adapting to nutritional stress, with \textit{H. lanatus} initially investing more in roots, while \textit{D. flexuosa} adjusts its allocation over time.

\begin{figure}
    \centering
    \includegraphics[scale=0.5]{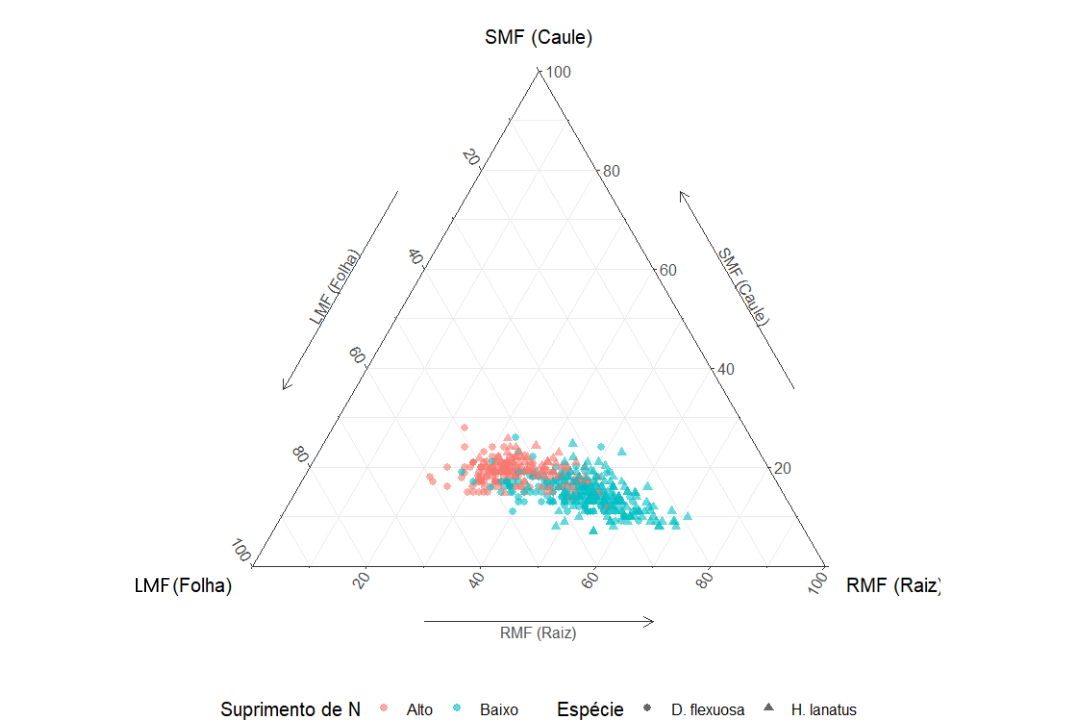}
    \caption{Ternary plot representing the proportions of biomass allocated to leaves (LMF), stems (SMF), and roots (RMF) for the species Deschampsia flexuosa (circles) and Holcus lanatus (triangles), cultivated under two nitrogen-supply levels (high and low).}
    \label{fig:ternario}
\end{figure}

Figure~\ref{fig:ternario} presents a ternary plot summarizing proportional biomass allocation among leaves (LMF), stems (SMF), and roots (RMF) for different combinations of species and nitrogen supply. Each point in the plot represents an individual, whose position is determined by the relative proportion of biomass allocated to each of the three compartments, summing to 100\%. The data refer to the species \textit{Deschampsia flexuosa} (circles) and \textit{Holcus lanatus} (triangles), cultivated under two nitrogen-supply levels: high (pink points) and low (blue points).

The points cluster in distinct regions of the plot, reflecting marked differences in biomass-allocation strategies across species and treatments. The species \textit{H. lanatus}, especially under low nitrogen supply, concentrates in the lower-right region of the triangle, suggesting greater allocation of biomass to roots (RMF). In contrast, \textit{D. flexuosa} tends to concentrate closer to the center and the left side of the plot, indicating a more balanced biomass distribution, with a higher proportion allocated to leaves (LMF) and lower relative investment in roots.

In addition, the effect of nitrogen supply is evident: individuals grown under low nitrogen (blue points) tend to shift toward regions with higher RMF, regardless of species, while individuals under high nitrogen (pink points) concentrate in regions with a more homogeneous distribution among leaves, stems, and roots. These patterns reinforce the joint influence of species identity and environmental conditions in shaping biomass-allocation strategies, highlighting the interaction between genotype and environment in plant growth.

\subsection{Model fitting}
Table~\ref{tab:coeficientes_dirichlet} presents the estimated coefficients of the Dirichlet regression model for the proportions of biomass allocated to stems (SMF) and roots (RMF), as well as for the precision model, using the specification given in \eqref{regressao-especificada}.

We used the same model specification as in \cite{douma2019analysing}, in which the authors considered complex interactions and variable dispersion. For selecting interactions and covariates, the authors relied on the Akaike Information Criterion (AIC).

\begin{table}[ht]
\centering
\caption{Parameter estimates and standard errors in the Dirichlet regression model for biomass proportions (SMF and RMF).}
\label{tab:biomassa_formatada}
\begin{tabular}{clccc}
\hline
Submodel & Covariate & Estimate & Std. Error & Exp(estim) \\
\hline
$\mu_{SMF}$ & Intercept & -1,1110 & 0,1532 & 0,329 \\
(Stem)      & Time$^2$  & -0,0200 & 0,0139 & 0,980 \\
            & $\log$(Total Biomass) & 0,0426 & 0,0314 & 1,044 \\
            & Species: \textit{H. lanatus} & 0,2140 & 0,0557 & 1,239 \\
            & Treatment: Low N & 0,0547 & 0,0369 & 1,056 \\
            & Time & 0,0273 & 0,0429 & 1,028 \\
            & Species $\times$ Low N & -0,1616 & 0,0628 & 0,851 \\
            & Species $\times$ Time & -0,0163 & 0,0616 & 0,984 \\
            & Low N $\times$ Time & -0,0230 & 0,0336 & 0,977 \\
            & Sp. $\times$ N $\times$ Time & 0,1363 & 0,0657 & 1,146 \\
\hline
$\mu_{RMF}$ & Intercept & -1,0179 & 0,1192 & 0,361 \\
(Root)      & Time$^2$  & -0,0280 & 0,0103 & 0,972 \\
            & $\log$(Total Biomass) & 0,1412 & 0,0243 & 1,152 \\
            & Species: \textit{H. lanatus} & 0,3442 & 0,0441 & 1,411 \\
            & Treatment: Low N & 0,6648 & 0,0286 & 1,944 \\
            & Time & -0,1293 & 0,0338 & 0,879 \\
            & Species $\times$ Low N & -0,0117 & 0,0485 & 0,988 \\
            & Species $\times$ Time & -0,0407 & 0,0495 & 0,960 \\
            & Low N $\times$ Time & 0,2590 & 0,0263 & 1,296 \\
            & Sp. $\times$ N $\times$ Time & -0,2430 & 0,0511 & 0,784 \\
\hline
$\phi$      & Intercept & 5,4220 & 0,0815 & 226,331 \\
(Precision) & Species: \textit{H. lanatus} & -0,2625 & 0,0931 & 0,769 \\
            & Time & 0,1180 & 0,0596 & 1,125 \\
            & Treatment: Low N & -0,4637 & 0,0976 & 0,629 \\
            & Species $\times$ Time & 0,2910 & 0,0933 & 1,338 \\
\hline
\end{tabular}
\label{tab:coeficientes_dirichlet}
\end{table}

Since we are in a misspecification scenario, it does not make sense to interpret the parameters, as this could lead to misleading conclusions about the effects of the variables. Moreover, some complex interactions do not allow for easy interpretability. Therefore, this is essentially a setting restricted to prediction.

\begin{figure}
    \centering
    \includegraphics[scale=0.5]{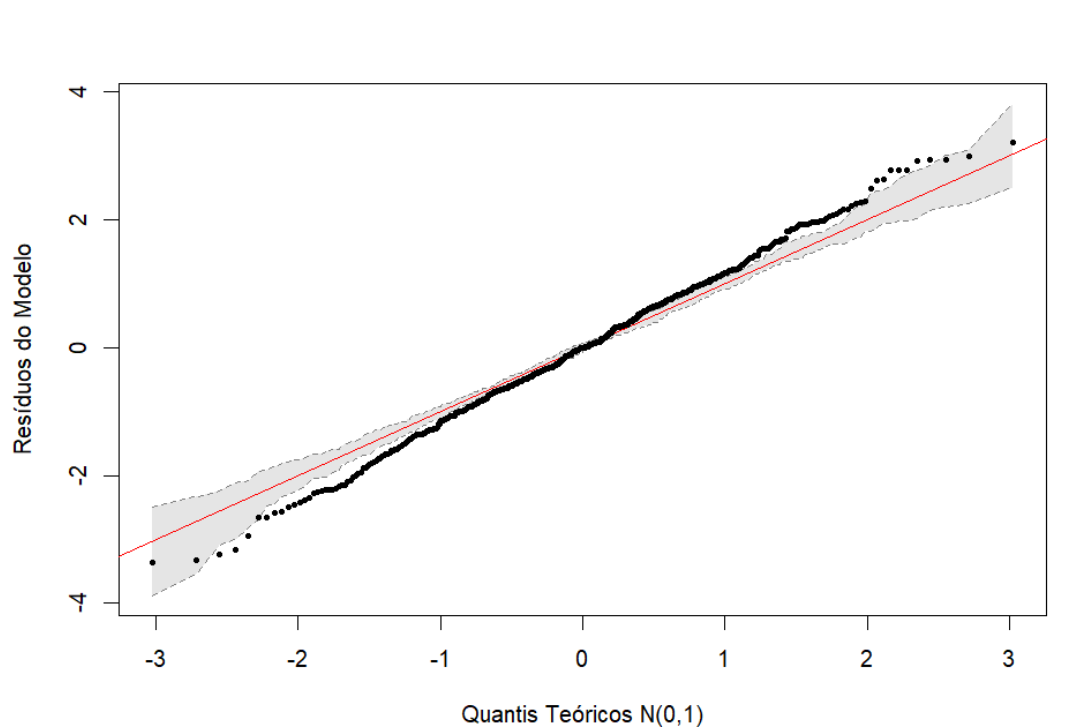}
    \caption{Normal probability plot with simulated envelope for the residuals.}
    \label{fig:qqplot-biomass}
\end{figure}
Finally, to verify whether the model is indeed misspecified, we used the normal probability plot with a simulated envelope for the residuals in Figure \ref{fig:qqplot-biomass}. Considering the final model, the plot suggests inadequacy of the Dirichlet regression model for the proportion of biomass allocated to each structure, since there are numerous points outside the envelope. Next, we evaluate this application in predictive terms under misspecification.

\subsection{Predictive Analysis}
Moreover, in order to assess the performance of the methods developed throughout this work, we analyzed the predictive sets generated from the real dataset. This step aims not only to check predictive accuracy, but above all to understand how the proposed approaches behave given the inherent complexity of proportional and correlated data, as in the case of biomass allocation among leaves, stems, and roots.

Table \ref{tab:preditivos_formatado} presents the predictive performance of the evaluated methods for the biomass-allocation application, in terms of empirical coverage and mean relative area of the prediction regions. Unlike the previous application, this is a deliberately misspecified scenario, in which inferential interpretation of the model parameters is not appropriate and the analysis focuses exclusively on the predictive quality of the constructed sets.

The quantile method yields an empirical coverage of 92,2\%, slightly above the nominal level, associated with an intermediate relative area. This result suggests that, even under misspecification, the procedure maintains reasonable coverage control, albeit at the cost of regions that do not fully exploit the simplex geometry and the inherent dependence among biomass proportions.

The HDR-aprox method again appears more conservative, achieving the highest empirical coverage (94,8\%), along with the largest mean relative area among the considered methods. This behavior reflects the procedure’s sensitivity to model misspecification, leading to wider predictive regions as a way to compensate for additional uncertainty in the estimated conditional distribution.

In contrast, HDR-aprox-grid attains an empirical coverage of 91,2\%, quite close to the nominal level, with the smallest mean relative area observed. This result shows that, even in a context where the underlying model is inadequate, combining the HDR approximation with verification on a physical grid can produce more concentrated and efficient predictive regions, without a relevant loss in coverage control.

Taken together, the results reinforce the robustness of SCP-based methods to model misspecification. In particular, the performance of HDR-aprox-grid highlights its ability to adapt to complex and non-ideal settings, offering a more favorable compromise between probabilistic accuracy and geometric parsimony. In the biomass-allocation context, where the relationships among compartments are strongly interdependent and shaped by environmental and ontogenetic factors, this method appears especially suitable for predictive purposes.

\begin{table}[ht]
\centering
\caption{Performance of predictive methods in terms of empirical coverage and mean relative area.}
\label{tab:preditivos_formatado}
\begin{tabular}{lcc}
\hline
\textbf{Method} & \textbf{Empirical Coverage (\%)} & \textbf{Relative area} \\ \hline
Quantile       & 92,2                           & 0,0423               \\
HDR-aprox      & 94,8                           & 0,0613               \\
HDR-aprox-grid & 91,2                           & 0,0370               \\ \hline
\end{tabular}
\end{table}

To better understand the shape and behavior of the predictive sets obtained by the different methods, we conducted a visual analysis of the prediction regions in individual examples from the test set. This visualization is particularly useful in compositional-data contexts, as it allows one to directly observe the geometry of the regions in the ternary simplex space, as well as the relative position of the observations with respect to the generated sets.

\begin{figure}
    \centering
    \includegraphics[scale = 0.7]{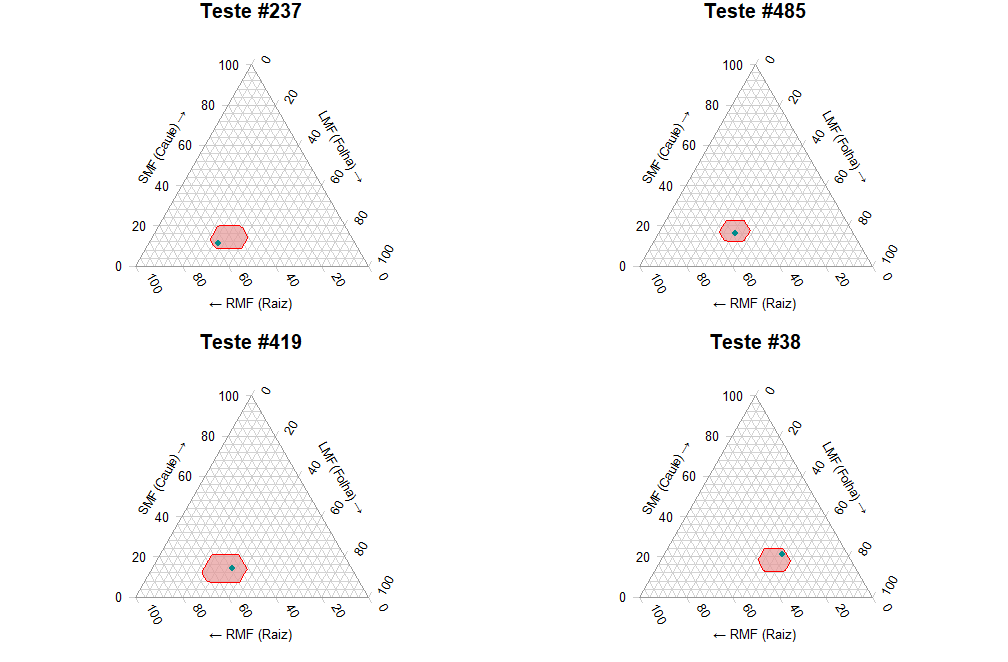}
    \caption{Prediction regions obtained by the method based on quantile residuals for different compositions. The red areas in the ternary plots represent the prediction regions. The green points correspond to the true observations in each test.}
    \label{fig:quantile_regions}
\end{figure}

The plots in Figure~\ref{fig:quantile_regions} show that the method based on quantile residuals produces regions built from independent marginal bounds for each component of the composition. Although this approach is simple and computationally efficient, it may generate predictive regions that do not respect the joint structure of the variables, which can lead to the inclusion of infeasible compositions.

\begin{figure}[ht]
    \centering
    \includegraphics[scale = 0.7]{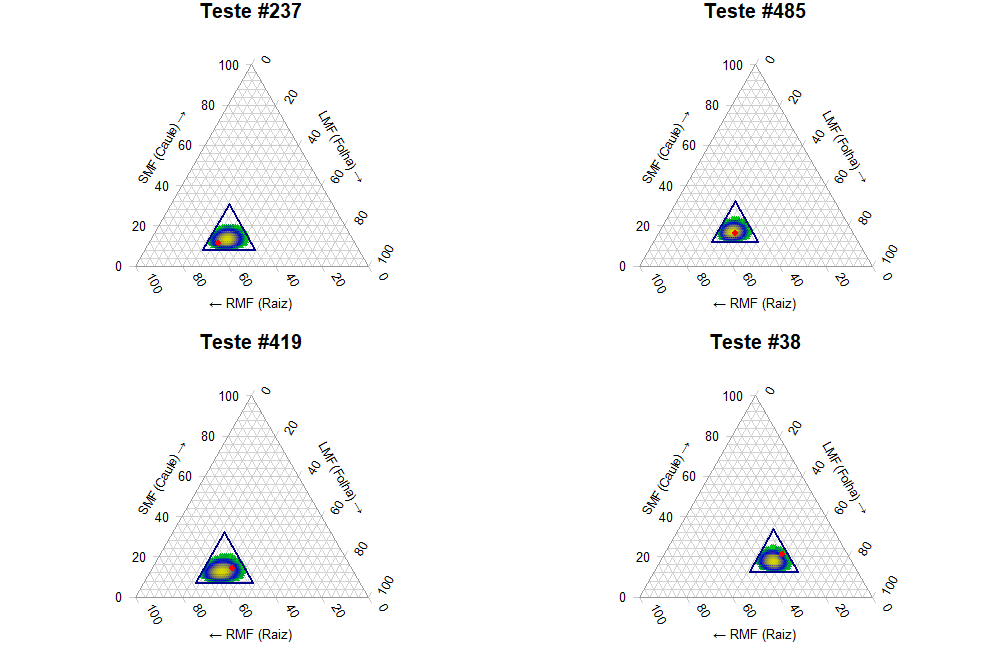}
    \caption{Prediction regions obtained by the HDR-aprox and HDR-aprox-grid methods for different compositions. The colored areas in the ternary plots indicate regions of higher predictive density, where future values are more likely to occur. The red points represent the true observations for each test. HDR-aprox is represented by the triangle that encloses the HDR approximation via a \textit{grid} (HDR-aprox-grid), shown by the color gradient from blue to yellow, indicating higher probabilities as the color approaches yellow.}
    \label{fig:hdr_regions}
\end{figure}

In contrast, the results in Figure~\ref{fig:hdr_regions} show that the HDR-aprox-grid method generates regions that are better adapted to the shape of the fitted Dirichlet distribution, resulting in sets with smooth shapes aligned with the highest-density areas.

\section{Discussion}
\label{sec:conclusions}
This work proposed three approaches for constructing valid prediction regions in Dirichlet regression models, with a focus on compositional data. They were compared through simulation studies and evaluated in two applications, one related to sleep-stage data and the other in the context of plant biomass allocation. The proposed approaches were: the method based on quantile residuals, the high-density region approximation approach (HDR-aprox), and its grid-based extension, HDR-aprox-grid.

The experiments revealed important differences among the approaches. The quantile residual method exhibited very stable performance, producing relatively compact regions and requiring little computation time, which makes it attractive when simplicity and efficiency are desired. Still, because it relies on an essentially marginal construction, its behavior can become more sensitive in settings where the distribution over the simplex is more irregular, which limits its robustness in some scenarios.

The HDR-aprox approach, in turn, proved to be systematically more conservative. By approximating the high-density region through the floor polytope, it tends to incorporate an additional safety margin, resulting in larger predictive sets. This pattern suggests greater robustness, especially when the data-generating mechanism departs from the fitted model, but at the cost of regions that are less informative from a geometric standpoint.

In this context, HDR-aprox-grid stood out as the most efficient alternative in terms of set size while preserving adequate coverage. By restricting the search to the interior of the polytope and discretizing the region in a way that is compatible with the geometry of the predicted density, the method reduces the unnecessary expansion observed in HDR-aprox and, in general, also improves upon the quantile procedure in relative area. On the other hand, this improvement comes with higher computational cost due to grid evaluation. Despite this, the comparison with naive discretization over the entire simplex shows that exploiting the floor polytope is crucial to making grid-based construction feasible in higher dimensions, avoiding the explosive growth in the cost of exhaustive search.

In summary, the experiments indicate that the quantile-residual method provides a fast and stable solution, HDR-aprox prioritizes robustness through increased conservatism, and HDR-aprox-grid offers a particularly attractive compromise when one wishes to reduce set size while maintaining predictive validity.

In addition, graphical representations in ternary diagrams made it possible to observe how the predictive regions behave geometrically. This provided visual evidence that HDR-based regions more faithfully follow the contours of the predicted densities, whereas the quantile-based method produces more symmetric and regular regions that are less aligned with the true shape of the distributions.

As perspectives for future work, several promising directions may complement and extend the results obtained here. First, there is the development of methods for small samples. Although the full conformal method is a natural alternative, the grid-based search combined with repeated model fitting makes its use infeasible in this setting. Moreover, a similar study could be developed for compositional samples that contain zeros. Finally, future work could investigate conformal prediction for compositional data using transformed models which, unlike the Dirichlet distribution, allow for the capture of positive correlations, for instance.

\bibliographystyle{natbib}
\bibliography{Resumo/referencias}

@book{vovk2005algorithmic,
  title={Algorithmic learning in a random world},
  author={Vovk, Vladimir and Gammerman, Alexander and Shafer, Glenn},
  volume={29},
  year={2005},
  publisher={Springer}
}

@article{shafer2008tutorial,
  title={A tutorial on conformal prediction},
  author={Shafer, Glenn and Vovk, Vladimir},
  journal={Journal of Machine Learning Research},
  volume={9},
  number={3},
  year={2008}
}

@inproceedings{papadopoulos2008normalized,
  title={Normalized nonconformity measures for regression conformal prediction},
  author={Papadopoulos, Harris and Gammerman, Alex and Vovk, Volodya},
  booktitle={Proceedings of the IASTED International Conference on Artificial Intelligence and Applications (AIA 2008)},
  pages={64--69},
  year={2008}
}

@article{lei2018distribution,
  title={Distribution-free predictive inference for regression},
  author={Lei, Jing and G’Sell, Max and Rinaldo, Alessandro and Tibshirani, Ryan J and Wasserman, Larry},
  journal={Journal of the American Statistical Association},
  volume={113},
  number={523},
  pages={1094--1111},
  year={2018},
  publisher={Taylor \& Francis}
}

@article{tibshirani2019conformal,
  title={Conformal prediction under covariate shift},
  author={Tibshirani, Ryan J and Foygel Barber, Rina and Candes, Emmanuel and Ramdas, Aaditya},
  journal={Advances in Neural Information Processing Systems},
  volume={32},
  year={2019}
}

@article{romano2019conformalized,
  title={Conformalized quantile regression},
  author={Romano, Yaniv and Patterson, Evan and Candes, Emmanuel},
  journal={Advances in Neural Information Processing Systems},
  volume={32},
  year={2019}
}

@article{barber2021predictive,
  title={Predictive inference with the jackknife+},
  author={Barber, Rina Foygel and Candes, Emmanuel J and Ramdas, Aaditya and Tibshirani, Ryan J},
  journal={The Annals of Statistics},
  volume={49},
  number={1},
  pages={486--507},
  year={2021},
  publisher={Institute of Mathematical Statistics}
}

@article{angelopoulos2021gentle,
  title={A gentle introduction to conformal prediction and distribution-free uncertainty quantification},
  author={Angelopoulos, Anastasios N and Bates, Stephen},
  journal={arXiv preprint arXiv:2107.07511},
  year={2021}
}

@inproceedings{kato2023review,
  title={A review of nonconformity measures for conformal prediction in regression},
  author={Kato, Yuko and Tax, David MJ and Loog, Marco},
  booktitle={Conformal and probabilistic prediction with applications},
  pages={369--383},
  year={2023},
  publisher={PMLR}
}

@article{barber2023conformal,
  title={Conformal prediction beyond exchangeability},
  author={Barber, Rina Foygel and Candes, Emmanuel J and Ramdas},
  journal={The Annals of Statistics},
  volume={51},
  number={2},
  pages={816--845},
  year={2023},
  publisher={Institute of Mathematical Statistics}
}

@inproceedings{bostrom2020mondrian,
  title={Mondrian conformal regressors},
  author={Bostr{\"o}m, Henrik and Johansson, Ulf},
  booktitle={Conformal and probabilistic prediction and applications},
  pages={114--133},
  year={2020},
  organization={PMLR}
}

@article{chernozhukov2021distributional,
  title={Distributional conformal prediction},
  author={Chernozhukov, Victor and W{\"u}thrich, Kaspar and Zhu, Yinchu},
  journal={Proceedings of the National Academy of Sciences},
  volume={118},
  number={48},
  pages={e2107794118},
  year={2021},
  publisher={National Academy of Sciences}
}

@inproceedings{vovk2012conditional,
  title={Conditional validity of inductive conformal predictors},
  author={Vovk, Vladimir},
  booktitle={Asian Conference on Machine Learning},
  pages={475--490},
  year={2012},
  organization={PMLR}
}

@article{dewolf2025conditional,
  title={Conditional validity of heteroskedastic conformal regression},
  author={Dewolf, Nicolas and De Baets, Bernard and Waegeman, Willem},
  journal={Information and Inference: A Journal of the IMA},
  volume={14},
  number={2},
  pages={iaaf013},
  year={2025},
  publisher={Oxford University Press}
}

@article{wu2025conformalized,
  title={Conformalized Regression for Continuous Bounded Outcomes},
  author={Wu, Zhanli and Leisen, Fabrizio and Rubio, F Javier},
  journal={arXiv preprint arXiv:2507.14023},
  year={2025}
}

@book{Izbicki2025,
  author    = {Rafael Izbicki},
  title     = {Machine Learning Beyond Point Predictions: Uncertainty Quantification},
  edition   = {1st},
  year      = {2025},
  pages     = {260},
  isbn      = {978-65-01-20272-3}
}

@article{cabezas2025regression,
  title = {Regression trees for fast and adaptive prediction intervals},
  journal = {Information Sciences},
  volume = {686},
  pages = {121369},
  year = {2025},
  issn = {0020-0255},
  doi = {https://doi.org/10.1016/j.ins.2024.121369},
  url = {https://www.sciencedirect.com/science/article/pii/S0020025524012830},
  author = {Luben M.C. Cabezas and Mateus P. Otto and Rafael Izbicki and Rafael B. Stern}
}

@article{aitchison1982statistical,
  title={The statistical analysis of compositional data},
  author={Aitchison, John},
  journal={Journal of the Royal Statistical Society: Series B (Methodological)},
  volume={44},
  number={2},
  pages={139--160},
  year={1982},
  publisher={Wiley Online Library}
}

@article{barndorff1991some,
  title={Some parametric models on the simplex},
  author={Barndorff-Nielsen, Ole E and J{\o}rgensen, Bent},
  journal={Journal of Multivariate Analysis},
  volume={39},
  number={1},
  pages={106--116},
  year={1991},
  publisher={Elsevier}
}

@article{hijazi2009modelling,
  title={Modelling compositional data using Dirichlet regression models},
  author={Hijazi, Rafiq H and Jernigan, Robert W},
  journal={Journal of Applied Probability \& Statistics},
  volume={4},
  number={1},
  pages={77--91},
  year={2009}
}

@techreport{maier2014dirichletreg,
  author      = {Maier, Marco J.},
  title       = {DirichletReg: Dirichlet Regression for Compositional Data in R},
  institution = {Department of Statistics and Mathematics},
  series      = {Research Report Series},
  number      = {125},
  year        = {2014}
}

@article{alenazi2023review,
  title={A review of compositional data analysis and recent advances},
  author={Alenazi, Abdulaziz},
  journal={Communications in Statistics-Theory and Methods},
  volume={52},
  number={16},
  pages={5535--5567},
  year={2023},
  publisher={Taylor \& Francis}
}

@article{greenacre2021compositional,
  title={Compositional data analysis},
  author={Greenacre, Michael},
  journal={Annual Review of Statistics and its Application},
  volume={8},
  number={1},
  pages={271--299},
  year={2021},
  publisher={Annual Reviews}
}

@article{zhang2025compositional,
  title={A compositional approach to the analysis of white blood cell counts for early COVID-19 detection},
  author={Zhang, Zhilong and Graffelman, Jan and Dorn, Marcio},
  journal={medRxiv},
  pages={2025--05},
  year={2025},
  publisher={Cold Spring Harbor Laboratory Press}
}

@article{ferrari2004beta,
  author={Ferrari, S. and Cribari-Neto, F.},
  title={Beta regression for modelling rates and proportions},
  journal={Journal of Applied Statistics},
  volume={31},
  number={7},
  pages={799--815},
  year={2004},
  publisher={Taylor \& Francis}
}

@article{pereira2019quantile,
  title={On quantile residuals in beta regression},
  author={Pereira, G. H. A.},
  journal={Communications in Statistics-Simulation and Computation},
  volume={48},
  number={1},
  pages={302--316},
  year={2019},
  publisher={Taylor \& Francis}
}

@article{dunn1996randomized,
  title={Randomized quantile residuals},
  author={Dunn, Peter K and Smyth, Gordon K},
  journal={Journal of Computational and Graphical Statistics},
  volume={5},
  number={3},
  pages={236--244},
  year={1996},
  publisher={Taylor \& Francis}
}

@article{espinheira2014bootstrap,
  title={Bootstrap prediction intervals in beta regressions},
  author={Espinheira, Patr{\'\i}cia L and Ferrari, Silvia LP and Cribari-Neto, Francisco},
  journal={Computational Statistics},
  volume={29},
  pages={1263--1277},
  year={2014},
  publisher={Springer}
}

@article{cribari2021resampling,
  title={Resampling-based prediction intervals in beta regressions under correct and incorrect model specification},
  author={Cribari-Neto, Francisco and Lima, F{\'a}bio P},
  journal={Communications in Statistics-Simulation and Computation},
  volume={50},
  number={5},
  pages={1398--1416},
  year={2021},
  publisher={Taylor \& Francis}
}

@article{pereira2024class,
  title={A class of bootstrap based residuals for compositional data},
  author={Pereira, Gustavo HA and Cai, Jianwen},
  journal={arXiv preprint arXiv:2403.13544},
  year={2024}
}

@article{ghorbani2019mahalanobis,
  title={Mahalanobis distance and its application for detecting multivariate outliers},
  author={Ghorbani, Hamid},
  journal={Facta Universitatis, Series: Mathematics and Informatics},
  pages={583--595},
  year={2019}
}

@book{hosmer2013applied,
  title={Applied logistic regression},
  author={Hosmer Jr, David W and Lemeshow, Stanley and Sturdivant, Rodney X},
  year={2013},
  publisher={John Wiley \& Sons}
}

@article{hyndman1996computing,
  title={Computing and graphing highest density regions},
  author={Hyndman, Rob J},
  journal={The American Statistician},
  volume={50},
  number={2},
  pages={120--126},
  year={1996},
  publisher={Taylor \& Francis}
}

@article{douma2019analysing,
  title={Analysing continuous proportions in ecology and evolution: A practical introduction to beta and Dirichlet regression},
  author={Douma, Jacob C and Weedon, James T},
  journal={Methods in Ecology and Evolution},
  volume={10},
  number={9},
  pages={1412--1430},
  year={2019},
  publisher={Wiley Online Library}
}

@article{ritmala2015sleep,
  title={Sleep and nursing care activities in an intensive care unit},
  author={Ritmala-Castren, Marita and Virtanen, Irina and Leivo, Sanna and Kaukonen, Kirsi-Maija and Leino-Kilpi, Helena},
  journal={Nursing \& health sciences},
  volume={17},
  number={3},
  pages={354--361},
  year={2015},
  publisher={Wiley Online Library}
}

@article{hussain2022quantitative,
  title={Quantitative evaluation of EEG-biomarkers for prediction of sleep stages},
  author={Hussain, Iqram and Hossain, Md Azam and Jany, Rafsan and Bari, Md Abdul and Uddin, Musfik and Kamal, Abu Raihan Mostafa and Ku, Yunseo and Kim, Jik-Soo},
  journal={Sensors},
  volume={22},
  number={8},
  pages={3079},
  year={2022},
  publisher={MDPI}
}

@article{tempesta2018sleep,
  title={Sleep and emotional processing},
  author={Tempesta, Daniela and Socci, Valentina and De Gennaro, Luigi and Ferrara, Michele},
  journal={Sleep medicine reviews},
  volume={40},
  pages={183--195},
  year={2018},
  publisher={Elsevier}
}

@article{gonzalez2019analysis,
  title={Analysis of sleep macrostructure in patients diagnosed with Parkinson’s disease},
  author={Gonz{\'a}lez-Naranjo, Justa Elizabeth and Alfonso-Alfonso, Maydelin and Grass-Fernandez, Daymet and Morales-Chac{\'o}n, Lilia Mar{\'\i}a and Pedroso-Ib{\'a}{\~n}ez, Iv{\'o}n and Ricardo-de la Fe, Yordanka and Padr{\'o}n-S{\'a}nchez, Arnoldo},
  journal={Behavioral Sciences},
  volume={9},
  number={1},
  pages={6},
  year={2019},
  publisher={MDPI}
}

@article{maski2021stability,
  title={Stability of nocturnal wake and sleep stages defines central nervous system disorders of hypersomnolence},
  author={Maski, Kiran P and Colclasure, Alicia and Little, Elaina and Steinhart, Erin and Scammell, Thomas E and Navidi, William and Diniz Behn, Cecilia},
  journal={Sleep},
  volume={44},
  number={7},
  pages={zsab021},
  year={2021},
  publisher={Oxford University Press US}
}

@article{veje2021sleep,
  title={Sleep architecture, obstructive sleep apnea and functional outcomes in adults with a history of Tick-borne encephalitis},
  author={Veje, Malin and Studahl, Marie and Thunstr{\"o}m, Erik and Stentoft, Erika and Nolskog, Peter and Celik, Yeliz and Peker, Y{\"u}ksel},
  journal={PLoS One},
  volume={16},
  number={2},
  pages={e0246767},
  year={2021},
  publisher={Public Library of Science San Francisco, CA USA}
}

@article{lemonte2019residuals,
  title = {On residuals in generalized {Johnson} {SB} regressions},
  author = {Lemonte, Artur J. and Moreno-Arenas, Germ{\'a}n},
  journal = {Applied Mathematical Modelling},
  volume = {67},
  pages = {62--73},
  year = {2019},
  publisher = {Elsevier},
  doi = {10.1016/j.apm.2018.10.012}
}

@article{feng2020comparison,
  title = {A comparison of residual diagnosis tools for diagnosing regression models for count data},
  author = {Feng, Cindy and Li, Longhai and Sadeghpour, Alireza},
  journal = {BMC Medical Research Methodology},
  volume = {20},
  number = {1},
  pages = {175},
  year = {2020},
  publisher = {BioMed Central},
  doi = {10.1186/s12874-020-01055-2}
}

\clearpage
\appendix
\renewcommand{\thesection}{Appendix~\Alph{section}} 

\thispagestyle{empty}
\section{Split Conformal algorithm for the methods}
\label{apendiceA}

\begin{algorithm}[ht] \small
\caption{Split Conformal Prediction Set for Dirichlet Regression}
\label{alg:dirichlet_split_conformal_en}

\KwIn{Dataset $\mathcal{D}=\{(\mathbf{x}_i,\mathbf{y}_i)\}_{i=1}^n$ with $\mathbf{y}_i\in\Delta^{D}$; index split into training $I_1$ and calibration $I_2$ (with $n_{\text{cal}}=|I_2|$); model specification in \eqref{regressao-especificada}; significance level $\alpha\in(0,1)$.}
\KwOut{Conformal prediction set $\mathcal{C}_{\text{split}}(\mathbf{x}_{n+1})$ for a new covariate $\mathbf{x}_{n+1}$.}

\BlankLine
\begin{enumerate}[leftmargin=*,label=\textbf{\arabic*.}]
  \item \textbf{Split:} Partition the data into training $I_1$ and calibration $I_2$.
  \item \textbf{Fit:} Using $I_1$, fit the Dirichlet regression in \eqref{regressao-especificada} and obtain the parameters \\ $\widehat{\bm{\mu}}_i=\widehat{\bm{\mu}}(\mathbf{x}_i)$ and $\widehat{\phi}_i=\widehat{\phi}(\mathbf{x}_i)$.
  \item \textbf{Calibration scores:} For each $i\in I_2$:
  \begin{enumerate}[leftmargin=*,label=(\alph*)]
    \item For each component $j=1,\dots,D$, compute
    \[
      U_{ij} \;=\; F\bigl(y_{ij};\, \mu_{ij},\, \phi_{i}\bigr),
    \]
    where $F$ is the marginal beta CDF induced by the fitted model at \\ $\mathbf{x}_i$ for component $j$.
    \item Transform $U_{ij}$ via the standard normal: $r^q_{ij}=\Phi^{-1}(U_{ij})$.
    \item Define the nonconformity score: $S_i=\max_{j}\bigl|r^q_{ij}\bigr|$.
  \end{enumerate}
  \item \textbf{Conformal quantile:} Let $S_{(1)}\le\cdots\le S_{(n_{\text{cal}})}$ be the ordered $\{S_i:i\in I_2\}$. Define
  \[
    q_{1-\alpha}=S_{\left(\left\lceil (n_{\text{cal}}+1)(1-\alpha)\right\rceil\right)}.
  \]
  \item \textbf{Prediction for $\mathbf{x}_{n+1}$:}
  \begin{enumerate}[leftmargin=*,label=(\alph*)]
    \item Compute $p_{\inf}=\Phi(-q_{1-\alpha})$ and $p_{\sup}=\Phi(q_{1-\alpha})$.
    \item Obtain $\widehat{\bm{\mu}}_{n+1}=\widehat{\bm{\mu}}(\mathbf{x}_{n+1})$ and $\widehat{\phi}_{n+1}=\widehat{\phi}(\mathbf{x}_{n+1})$. For each component $j$, \\ construct  the marginal interval
    \[
      I_j(\mathbf{x}_{n+1})=\Bigl[F_{\mathrm{Beta}_j}^{-1}(p_{\inf})\;,\;
                          F_{\mathrm{Beta}_j}^{-1}(p_{\sup})\Bigr].
    \]
    \item Return the Cartesian prediction set restricted to the simplex:
    \[
      \mathcal{C}_{\text{split}}(\mathbf{x}_{n+1})=\Bigl\{\, \mathbf{y}\in\Delta^{D}:\; y_j\in I_j(\mathbf{x}_{n+1})\;\; \forall\, j=1,\dots,D \Bigr\}.
    \]
  \end{enumerate}
\end{enumerate}
\end{algorithm}

\begin{algorithm}[ht] \small
\caption{Split Conformal Prediction Set with HDR Approximation}
\label{alg:dirichlet_triangular_hdr_en}

\KwIn{Dataset $\mathcal{D}=\{(\mathbf{x}_i,\mathbf{y}_i)\}_{i=1}^n$ with $\mathbf{y}_i\in\Delta^{D}$; index split into training $I_1$ and calibration $I_2$ (with $n_{\text{cal}}=|I_2|$); Dirichlet model specification $\lambda_j(\mathbf{x})=\phi(\mathbf{x})\mu_j(\mathbf{x})$; level $\alpha\in(0,1)$.}
\KwOut{Conformal prediction set $\mathcal{T}(\mathbf{x}_{n+1})$ for a new covariate $\mathbf{x}_{n+1}$.}

\BlankLine
\begin{enumerate}[leftmargin=*,label=\textbf{\arabic*.}]
  \item \textbf{Split:} Partition the data into training $I_1$ and calibration $I_2$.

  \item \textbf{Model fit:} Using $I_1$, fit the Dirichlet regression and obtain the predictors
  \[
    \widehat{\bm{\mu}}(\mathbf{x}),\qquad \widehat{\phi}(\mathbf{x}).
  \]

  \item \textbf{Calibration scores (likelihood score):} For each $i\in I_2$, define
  \[
  s_i \;=\; -\log f\bigl(\mathbf{y}_i \mid \mathbf{x}_i;\,\widehat{\lambda}(\mathbf{x}_i)\bigr)
       \;=\; -\log\Gamma(\widehat{\phi}_i) + \sum_{j=1}^{D}\log\Gamma(\widehat{\mu}_{ij}\widehat{\phi}_i)
             - \sum_{j=1}^{D}(\widehat{\mu}_{ij}\widehat{\phi}_i - 1)\log y_{ij}.
  \]

  \item \textbf{Conformal quantile:} Let $s_{(1)}\le\cdots\le s_{(n_{\text{cal}})}$ be the ordered $\{s_i:i\in I_2\}$. Define
  \[
    q_{1-\alpha} \;=\; s_{\,\left(\left\lceil (n_{\text{cal}}+1)(1-\alpha)\right\rceil\right)} .
  \]

  \item \textbf{Parameters at the test point $\mathbf{x}_{n+1}$:} Compute
  \[
    \widehat{\bm{\mu}}_\ast=\widehat{\bm{\mu}}(\mathbf{x}_{n+1}),\quad
    \widehat{\phi}_\ast=\widehat{\phi}(\mathbf{x}_{n+1}),
  \]
  \[
    t_{1-\alpha} \;=\; -q_{1-\alpha} - \log \Gamma(\widehat{\phi}_\ast) 
                 + \sum_{j=1}^{D}\log \Gamma(\widehat{\mu}_{\ast j}\widehat{\phi}_\ast),\qquad
    w_j \;=\; \widehat{\phi}_\ast \widehat{\mu}_{\ast j}-1 .
  \]

  \item \textbf{Minimum floors via 1D optimization (one $i$ at a time):}
  For each $i\in\{1,\ldots,D\}$, solve the convex problem
  \[
    \min_{y\in\mathbb{R}^D} y_i\ \ \ 
    \text{subject to}\ \ \sum_{j} y_j=1,\ \ \sum_{j} w_j \log y_j \ge t_{1-\alpha},\ \ y_j\ge 0,
  \]
  obtaining $\tau_i$. Proceed in two steps:
  \begin{enumerate}[leftmargin=*,label=(\alph*)]
    \item \textbf{Closed form (interior case):} If $w_j>0$ for all $j$, use the KKT conditions to obtain
    \[
      \theta = \frac{1}{\tfrac{w_i}{1+\rho}+\tfrac{W-w_i}{\rho}},\qquad
      \tau_i=\frac{\theta\,w_i}{1+\rho},\qquad W=\sum_{j=1}^D w_j,
    \]
    where $\rho>0$ is the unique root of the one-dimensional equation
    \[
      F_i(\rho)\;=\; w_i \log\rho + (W-w_i)\log(1+\rho) - W\log\!\bigl(w_i\rho+(W-w_i)(1+\rho)\bigr)
                     + \sum_{j=1}^D w_j \log w_j \;-\; t_{1-\alpha} \;=\; 0.
    \]
    \item \textbf{Fallback (boundary/negative case):} If any $w_j\le 0$ (or the search fails), set $\tau_i = 0$.
  \end{enumerate}

  \item \textbf{Approximate set:} Define the floor polytope
  \[
    \mathcal{T}(\mathbf{x}_{n+1}) \;=\; \Bigl\{\, \mathbf{y}\in\Delta^D:\ y_i \ge \tau_i,\ \ i=1,\ldots,D \Bigr\}.
  \]
  \item \textbf{Output:} Return $\mathcal{T}(\mathbf{x}_{n+1})$ as the prediction set for $\mathbf{y}_{n+1}$.
\end{enumerate}
\end{algorithm}


\end{document}